\documentclass[aps,twocolumn,pre,floatfix,superscriptaddress,showpacs,showkeys]{revtex4-1}

\usepackage{subfigure}
\usepackage{graphicx}
\usepackage{epstopdf}
\usepackage{amssymb}
\usepackage[usenames]{color}
\usepackage{epsfig}
\usepackage{amsmath}
\usepackage{times}
\usepackage{bm}
\usepackage[normalem]{ulem}
\emergencystretch=\maxdimen
\begin{document}

\title{Long-term prediction of chaotic systems with recurrent neural networks}

\author{Huawei Fan}
\affiliation{School of Electrical, Computer, and Energy Engineering, Arizona State University, Tempe, Arizona 85287, USA}
\affiliation{School of Physics and Information Technology, Shaanxi Normal University, Xi'an 710062, China}

\author{Junjie Jiang}
\affiliation{School of Electrical, Computer, and Energy Engineering, Arizona State University, Tempe, Arizona 85287, USA}

\author{Chun Zhang}
\affiliation{School of Electrical, Computer, and Energy Engineering, Arizona State University, Tempe, Arizona 85287, USA}
\affiliation{School of Physics and Information Technology, Shaanxi Normal University, Xi'an 710062, China}

\author{Xingang Wang} 
\affiliation{School of Physics and Information Technology, Shaanxi Normal University, Xi'an 710062, China}

\author{Ying-Cheng Lai} \email{Ying-Cheng.Lai@asu.edu}
\affiliation{School of Electrical, Computer, and Energy Engineering, Arizona State University, Tempe, Arizona 85287, USA}
\affiliation{Department of Physics, Arizona State University, Tempe, Arizona 85287, USA}

\begin{abstract}

Reservoir computing systems, a class of recurrent neural networks, have recently been exploited for model-free, data-based prediction of the state evolution of a variety of chaotic dynamical systems. The prediction horizon demonstrated has been about half dozen Lyapunov time. Is it possible to significantly extend the prediction time beyond what has been achieved so far? We articulate a scheme incorporating time-dependent but sparse data inputs into reservoir computing and demonstrate that such rare ``updates'' of the actual state practically enable an arbitrarily long prediction horizon for a variety of chaotic systems. A physical understanding based on the theory of temporal synchronization is developed. 

\end{abstract}
\date{\today }
\maketitle

A recently emerged interdisciplinary field is machine-learning based, 
model-free prediction of the state evolution of nonlinear/chaotic dynamical 
systems~\cite{HSRFG:2015,LBMUCJ:2017,PLHGO:2017,LPHGBO:2017,DBN:book,
PWFCHGO:2018,PHGLO:2018,Carroll:2018,NS:2018,ZP:2018,WYGZS:2019,JL:2019}. 
A paradigm that has been exploited is reservoir computing~\cite{Jaeger:2001,
MNM:2002,JH:2004,MJ:2013}, a class of recurrent neural networks. Starting from 
the same initial condition, a well-trained reservoir system can generate a 
trajectory that stays close to that of the target system for a finite amount 
of time, realizing short-term prediction. Because of the hallmark of chaos - 
sensitive dependence on initial conditions, the solution of the reservoir 
system will diverge from that of the original system exponentially. 
Nonetheless, if training is done properly so that the single-step error is 
orders-of-magnitude smaller than the oscillation range of the chaotic 
signal~\cite{JL:2019}, accurate prediction can be achieved in short time. 
So far, the prediction horizon achieved is about five or six Lyapunov 
time~\cite{PHGLO:2018}, where one Lyapunov time is the inverse of the maximum 
Lyapunov exponent. 


Is it possible to extend significantly the prediction horizon of reservoir 
computing? We provide an affirmative answer in this paper.
The key observation is that, after training, prediction is enabled because 
the neural network system can replicate the dynamical evolution of the target 
system (or synchronize with it) but only for a transient period of time. In 
the conventional scheme, data from the target system are used only during the 
training phase. A solution to extend the transient time is to provide some 
``update'' of the target system. We thus conceive the scenario where, after 
the initial training, infrequent or sparse updates in the form of new 
measurement or observation of the target system are available. We demonstrate 
that even rare updates of the actual state enable an arbitrarily long 
prediction horizon to be achieved for a variety of chaotic systems. 
Essentially, before the trajectories of the reservoir and original systems 
diverge substantially (e.g., about to exceed a predefined accuracy), we 
correct the state of the reservoir system with real measurement of duration 
as few as a single data point. We develop a physical understanding based on 
the theory of temporal synchronization. Practically, with rare data updates, 
the reservoir computing system can replicate the evolution of the original 
system within some desired accuracy for an arbitrarily long time, in spite of 
chaos. This will have applications in fields where chaos arises.

\begin{figure}[ht!]
\begin{center}
\includegraphics[width=\linewidth]{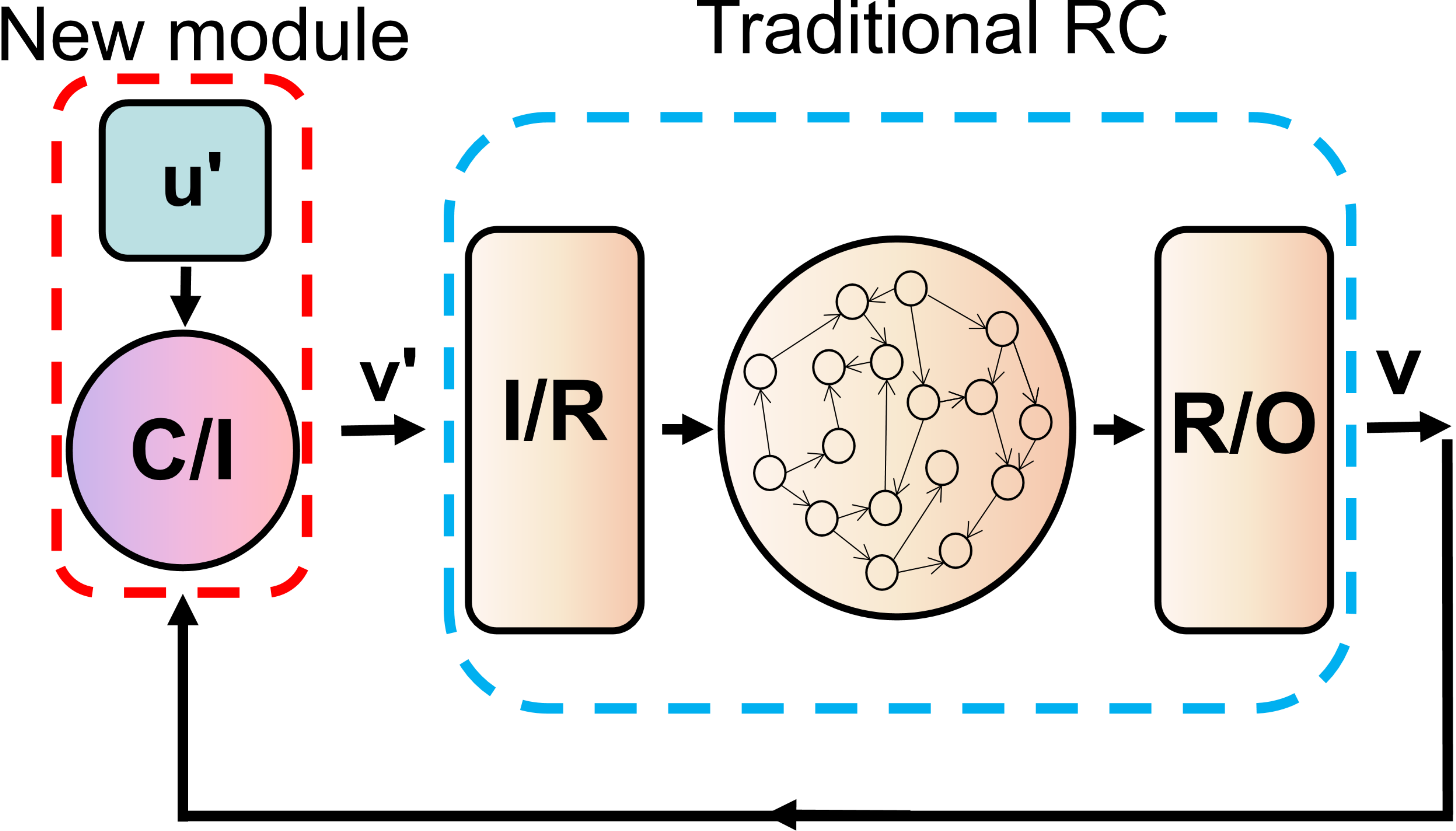}
\caption{Proposed reservoir computing (RC) system with rare state updates in 
the prediction phase, which is capable of generating arbitrarily long 
prediction of the state evolution of any chaotic system. The I/R, Reservoir, 
and R/O modules within the blue dashed box constitute the conventional 
reservoir computing system. Initial training with data from the target chaotic 
system is done in a conventional manner. The articulated scheme of rare state 
updates is represented by the C/I module inside the red dashed box, which 
couples sparse measurement data with the output of the system to generate 
updated inputs to the system.}
\label{fig:scheme}
\end{center}
\end{figure}

The basic working of reservoir computing can be described briefly as follows. 
Suppose time series data represented by a relatively low-dimensional data 
vector from the target system to be predicted are available. As shown in 
Fig.~\ref{fig:scheme}, one feeds the time series into the 
input-to-reservoir (I/R) module to generate a time-dependent data vector 
whose dimension is significantly larger than that of the original data vector. 
The high-dimensional vector is then sent to a complex network constituting the 
core of the reservoir, whose size matches the dimension of the vector. That 
is, there is a one-to-one correspondence between any component of 
the high-dimensional vector and a node in the network, and the data from a
component is fed into the corresponding node. The state of the reservoir
network is updated according to some nonlinear function, and the 
resulting state vector is sent to a reservoir-to-output (R/O) module whose
function is opposite to that of the I/R module, i.e., to convert the 
high-dimensional vector of the reservoir back into a vector of the same 
low-dimension as that of the original data from the target system. The 
reservoir network, once chosen, is fixed, so all parameters associated with 
it are hyperparameters. Training is done through a relatively small set of 
adjustable parameters associated with the R/O module, which can be tuned 
(or ``learned'') based on the available input data. During the training phase, 
the system is open as it requires input from the target system. After training,
one feeds the output from R/O directly into the I/R module, closing the system.
The system then evolves by itself. Some key numbers involved in reservoir 
computing for model-free prediction are as follows. For example, for the 
Kuramoto-Sivashinsky equation (KSE) with spatiotemporally chaotic 
solutions~\cite{PHGLO:2018}, the typical number of the spatial measurement 
sites (the dimension of the input vector) is 64 and the size of the reservoir 
network is about 5000. The number of parameters in the R/O module to be 
trained is $5000 \times 64$. 

Our proposed reservoir computing system functioning in the prediction phase 
leading to arbitrarily long prediction horizon is shown schematically in 
Fig.~\ref{fig:scheme}, where the three modules inside the blue dashed box 
represent the conventional system and the one inside the red dashed box is a 
new module incorporating rare state updates. The $I/R$ module 
is described by $\bm{W}_{in}$, a $D_{r}\times D_{in}$ random matrix that maps
a $D_{in}$-dimensional input vector $\bm{v}$ into a $D_r$-dimensional vector
$\bm{r}(t)$, where $D_{in} \ll D_r$. The elements of $\bm{W}_{in}$ are 
generated from a uniform distribution in $[-\sigma,\sigma]$. The reservoir is 
a complex network of $D_r$ nodes with average degree $d$, whose connecting 
topology is described by the $D_r\times D_r$ matrix $\bm{A}$. (For simplicity, 
we choose it to be a directed random network.) A recent work~\cite{JL:2019} 
has revealed that successful training and finite-time prediction can be 
achieved if the spectral radius of the network is in a finite range, which 
can be adjusted by properly normalizing the link weights in the network. The 
$R/I$ module is represented by a $D_{out}\times D_{r}$ matrix, whose elements 
are parameters determined through training~\cite{PLHGO:2017,PHGLO:2018}. 
Typically, we have $D_{out} = D_{in}$. Without any state update, in the 
prediction phase, the reservoir computing system is a self-evolving dynamical 
system described by 
$\bm{r}(t+\Delta t)=\tanh{[\bm{A}\cdot\bm{r}(t)+\bm{W}_{in}\cdot\bm{v}(t)]}$
and $\bm{v}(t+\Delta t)=\bm{W}_{out}\cdot\bm{f}[\bm{r}(t+\Delta t)]$,
where $\bm{f}(\bm{r})$ is the output function~\cite{PLHGO:2017,PHGLO:2018}: 
$f_i(\bm{r})=r_{i}$ and $f_i(\bm{r})=r^{2}_{i}$ for odd and even index $i$, 
($i=1,2,\ldots,D_{r}$), respectively. Associated with the dynamical evolution
of the reservoir system are two sets of dynamical variables: the 
high-dimensional reservoir state vector $\bm{r}$ and the (typically) 
low-dimensional output vector $\bm{v}$. 
The matrices $\bm{W}_{in}$ and $\bm{A}$ are pre-defined while $\bm{W}_{out}$ 
is determined by training during which $\bm{v}$ is replaced by the state 
vector $\bm{u}$ of the target system. After training is completed, to set 
the initial condition for the reservoir network, one approach is to continue 
to use $\bm{u}$ in place of $\bm{v}$ but only for a few time steps, after 
which the reservoir system executes natural dynamical evolution by itself. 

In the conventional scheme~\cite{HSRFG:2015,LBMUCJ:2017,PLHGO:2017,
LPHGBO:2017,DBN:book,PWFCHGO:2018,PHGLO:2018,Carroll:2018,NS:2018,
ZP:2018,WYGZS:2019,JL:2019}, after withdrawing the true state vector $\bm{u}$
so that the system is closed, real measurements are no longer used, resulting
in a relatively short prediction horizon for chaotic systems because of
the exponential divergence between the trajectories of the reservoir and
true systems. Our idea, as shown in Fig.~\ref{fig:scheme}, is to update the 
reservoir state with sparsely sampled real state vector $\bm{u}'$ before the 
divergence exceeds a pre-defined tolerance limit, i.e., the updates are needed 
only rarely. In particular, during the update, the input to the I/R module 
can be written as $\bm{v}'(t)=\bm{v}(t)+c[\bm{u}'(t)-\bm{v}(t)]$, where 
$\bm{u}'$ contains data at $t$ and $c$ is the coupling parameter. Most of the 
time during the evolution, we still have $\bm{v}'(t)=\bm{v}(t)$. Updating 
is effectively an on-off coupling process between the reservoir and 
the true systems, where the ``on'' phase is significantly more sparse than 
the ``off'' phase. Prediction can then be viewed as a  
synchronization process~\cite{LHO:2018,WYGZS:2019} between the two systems 
that are coupled but only intermittently~\cite{CQH:2009,LSCW:2018}. 

\begin{figure*}[t]
\centering
\includegraphics[width=\linewidth]{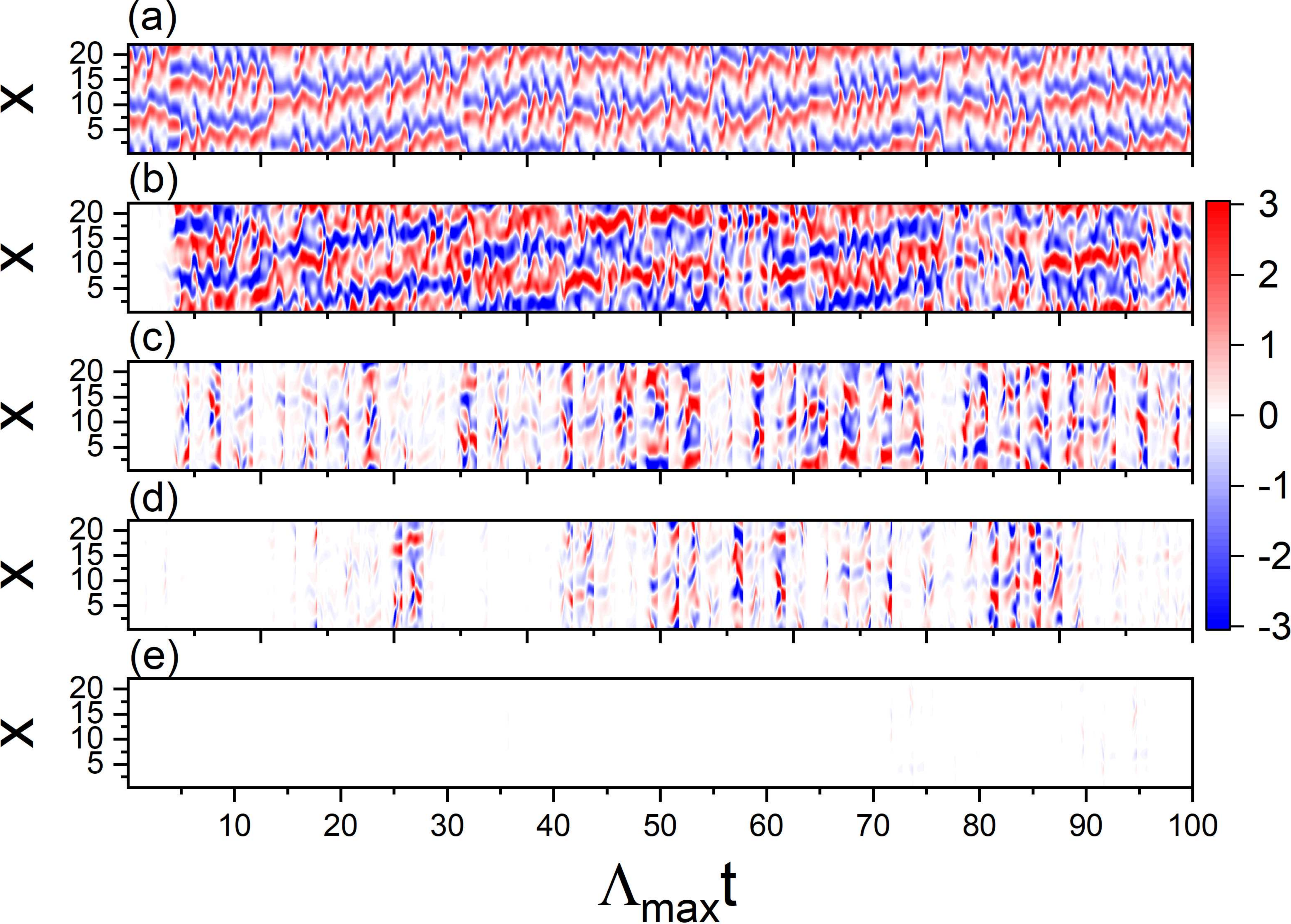}
\caption{ {\em Prediction of spatiotemporal chaotic solution of KSE}. 
The system size is $L=22$, the number of measurement points is $M = 64$, and
the value of the coupling parameter is $c = 1.0$. (a) True spatiotemporal 
evolution of the chaotic solution. (b) The difference (error) between the 
predicted and true solutions without any measurement updating, i.e., with 
updating period $T = \infty$. The prediction horizon is about five Lyapunov 
time. (c-e) The error with updating period $T = 240$, 160, and 80 time steps, 
corresponding to three, two, and one Lyapunov time, respectively, where the 
update consists of a single date point from the true system. In (e), error 
at all time is below the tolerance, signifying an arbitrarily long prediction 
horizon.}
\label{fig:KSE_1}
\end{figure*}

We test the predictive power of our proposed reservoir computing scheme 
with a large number of chaotic systems. Here we present two examples of 
high-dimensional chaotic systems: KSE and the complex Ginzburg-Landau
equation (cGLE). (Examples of a number of low-dimensional chaotic systems 
are presented in Appendices.) The KSE is 
$y_{t}+yy_{x}+y_{xx}+y_{xxxx}=0$, where $y(x,t)$ is a scalar field in the 
interval $x \in (0,L)$ with periodic boundaries. We divide the spatial domain
into $M$ uniform subintervals. Figure~\ref{fig:KSE_1}(a) shows a typical
spatiotemporal chaotic solution for $L=22$, where the numerical integration 
parameters are $M=64$ and $\Delta t=0.25$, and the maximum Lyapunov exponent 
is $\Lambda_{max}\approx 0.05$. Thus, approximately 80 time steps correspond
to one Lyapunov time. The $M$-dimensional data vector is fed into the reservoir
computing system with parameters $D_{in}=D_{out}=M$, $D_{r}=4992$, $\sigma=1$, 
$d=3$, and $\rho=0.1$. In addition, to avoid overfitting of $\bm{W}_{out}$ 
during the training process, we set the relevant bias 
parameter~\cite{PHGLO:2018} to be $\eta=1\times10^{-4}$. 
Figure~\ref{fig:KSE_1}(b) shows the 
difference between the evolution of the reservoir and true system, i.e., the
prediction error (color coded). The prediction horizon is about five Lyapunov
time. 

We now instigate rare updates with the true data $y_{act}(x,t)$, where the
coupling function is $y_{rc}'(x,t)=y_{rc}(x,t)+c[y_{act}(x,t)-y_{rc}(x,t)]$ 
if there is data update at the space-time point $(x,t)$, otherwise we have
$y_{rc}'(x,t)=y_{rc}(x,t)$, where $y_{rc}$ is the output of the reservoir 
system. We divide the output time series into equal time intervals, each of 
$T$ time steps. In one interval, the actual data points are available for 
consecutive $T_{0}$ steps, and there is no update for the remaining $T-T_{0}$ 
steps. In the space, we select $M_{c}$ uniform spatial points from the total 
$M$ measurement points. 
For illustration, we $T_{0}=1$ and 
$M_{c}=64$, i.e., each component of the input vector to the reservoir system 
(corresponding to a distinct measurement point in space) receives one true 
data point every $T$ time steps. Figure~\ref{fig:KSE_1}(c-e) show the 
prediction error for $T = 240$, 160, and 80 time steps, respectively, 
corresponding to three, two, and one Lyapunov time. For $T = 240$ 
[Fig.~\ref{fig:KSE_1}(c)], the error is reduced as compared the case of no 
data update [Fig.~\ref{fig:KSE_1}(b)] and exhibits intermittency,  
implying an intermittently synchronous behavior between the 
reservoir and the true system. In this case, the time interval in 
which the error is below the tolerance emerges intermittently in time.  
As the updating becomes twice as frequent, the small-error intervals are
generally enlarged, as shown in Fig.~\ref{fig:KSE_1}(d) for $T = 160$. 
Remarkable, for $T = 80$, the error is essentially zero in the whole time 
interval considered. 
This means that, insofar 
as a single true data point is used to update the reservoir system in every 
Lyapunov time, the prediction horizon becomes arbitrarily long! 

\begin{figure*}[t]
\centering
\includegraphics[width=0.9\linewidth]{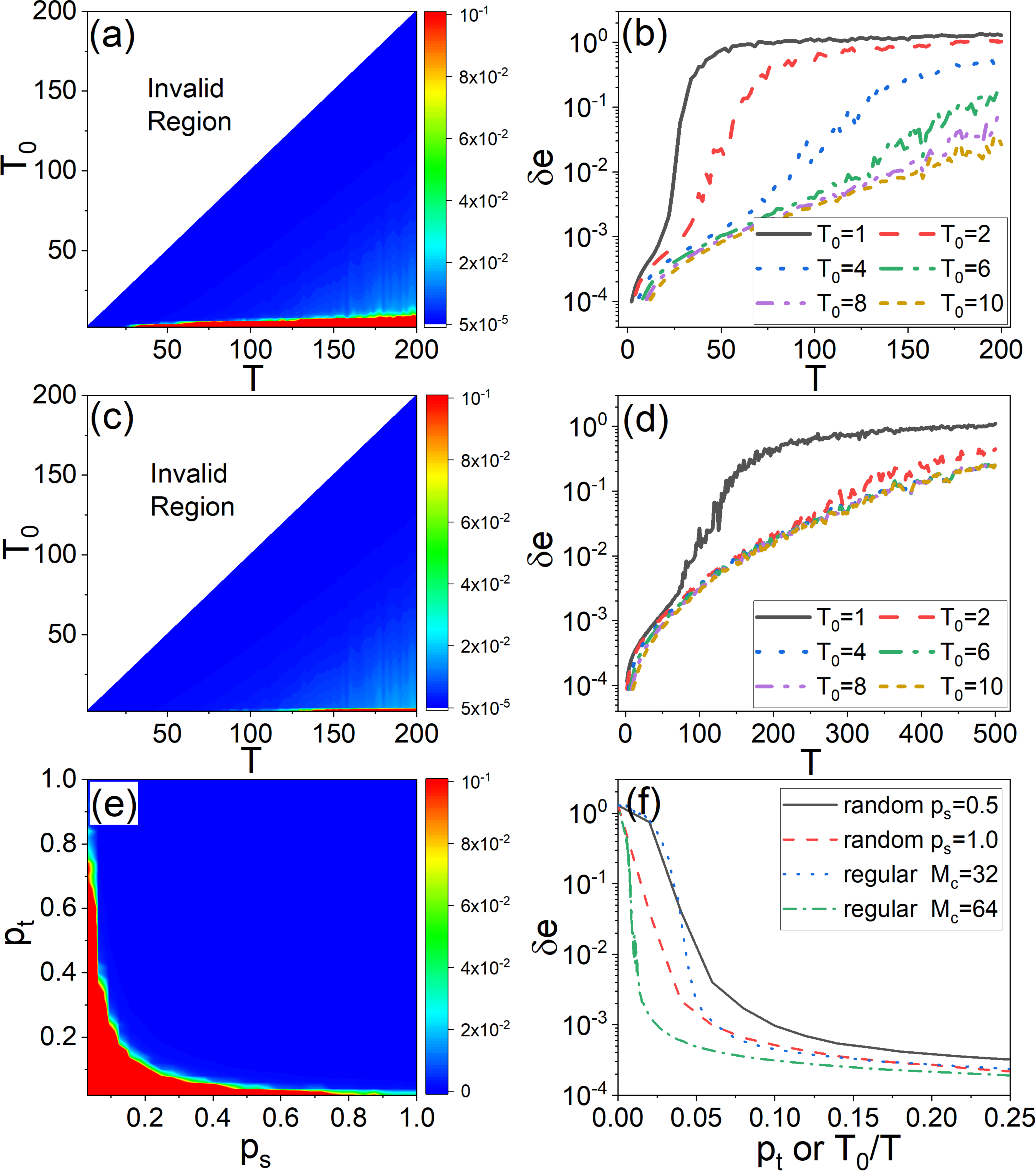}
\caption{ {\em Behaviors of prediction error for KSE}. 
(a) Prediction error in the parameter plane $(T_{0},T)$ for $M_c=32$ and 
$c=2.0$. (b) For $M_c=32$, prediction error $\delta e$ versus $T$ for 
different values of $T_{0}$. (c) Prediction error for $M_c=64$ and $c=1.0$. 
(d) For $M_c=64$, $\delta e$ versus $T$ for different values of $T_0$. 
(e) For random updating, prediction error in the parameter plane $(p_t,p_s)$ 
for $c=1.2$ and (f) $\delta e$ versus $p_t$ (or $T_0/T$).}
\label{fig:KSE_2}
\end{figure*}

To obtain a systematic picture of the prediction horizon, we calculate the 
average error $\delta e$ over a long time interval, e.g., about 100 Lyapunov
time, in the parameter plane $(T_{0}, T)$ (for $T_{0}\leq T$). 
Figure~\ref{fig:KSE_2}(a) shows, for $M_{c} = M/2 = 32$ (the number of spatial 
coupling channels) and $c=2.0$, the error behavior.
In most of the parameter region, the error is small: $\delta e<0.01$. Long
term prediction fails only for extremely rare updates, i.e., 
for small values of $T_{0}$ and large values of $T$. Fixing the value of 
$T_0$, we obtain the curves of $\delta e$ versus $T$. For $T_{0}=1$ (black 
line), $\delta e$ increases rapidly with $T$. As more data points are included
in each update, e.g., as the value of $T_0$ is increased from two to ten, 
$\delta e$ decreases gradually. Figure~\ref{fig:KSE_2}(c) shows that, for 
$M_c = M = 64$, the error remains small in all cases, giving rise to a
significantly augmented prediction horizon. Figure~\ref{fig:KSE_2}(d) shows
the prediction error for $T$ up to $500$ time steps. For a change from 
$T_{0}=1$ to $T_{0}=2$, error $\delta e$ decreases sharply. However, for
$T_{0}\geq 2$, the predictions error becomes saturated. Comparing 
Figs.~\ref{fig:KSE_2}(b) with \ref{fig:KSE_2}(d), we observe a dramatic 
reduction in $\delta e$ for $M_{c} = 64$. 

The cases considered so far are for rare but regular updates. What about
random updates? We examine the parameter plane 
$(p_{t},p_{s})$, where $p_{t}$ is the probability of update in each time 
step and $p_{s}$ is the probability for a spatial point to receive updates.
Figure~\ref{fig:KSE_2}(e) shows the average prediction error, which 
exhibits an approximately symmetric behavior with respect to $p_t$ and $p_s$. 
Figure~\ref{fig:KSE_2}(f) shows, for fixed $p_s=0.5$ and $p_s=1$ (corresponding 
to the cases of $M_c=32$ and $M_c=64$ with regular coupling, respectively), 
random updating yields somewhat larger errors than those with regular 
coupling.  

\begin{figure*}[ht!]
\centering
\includegraphics[width=\linewidth]{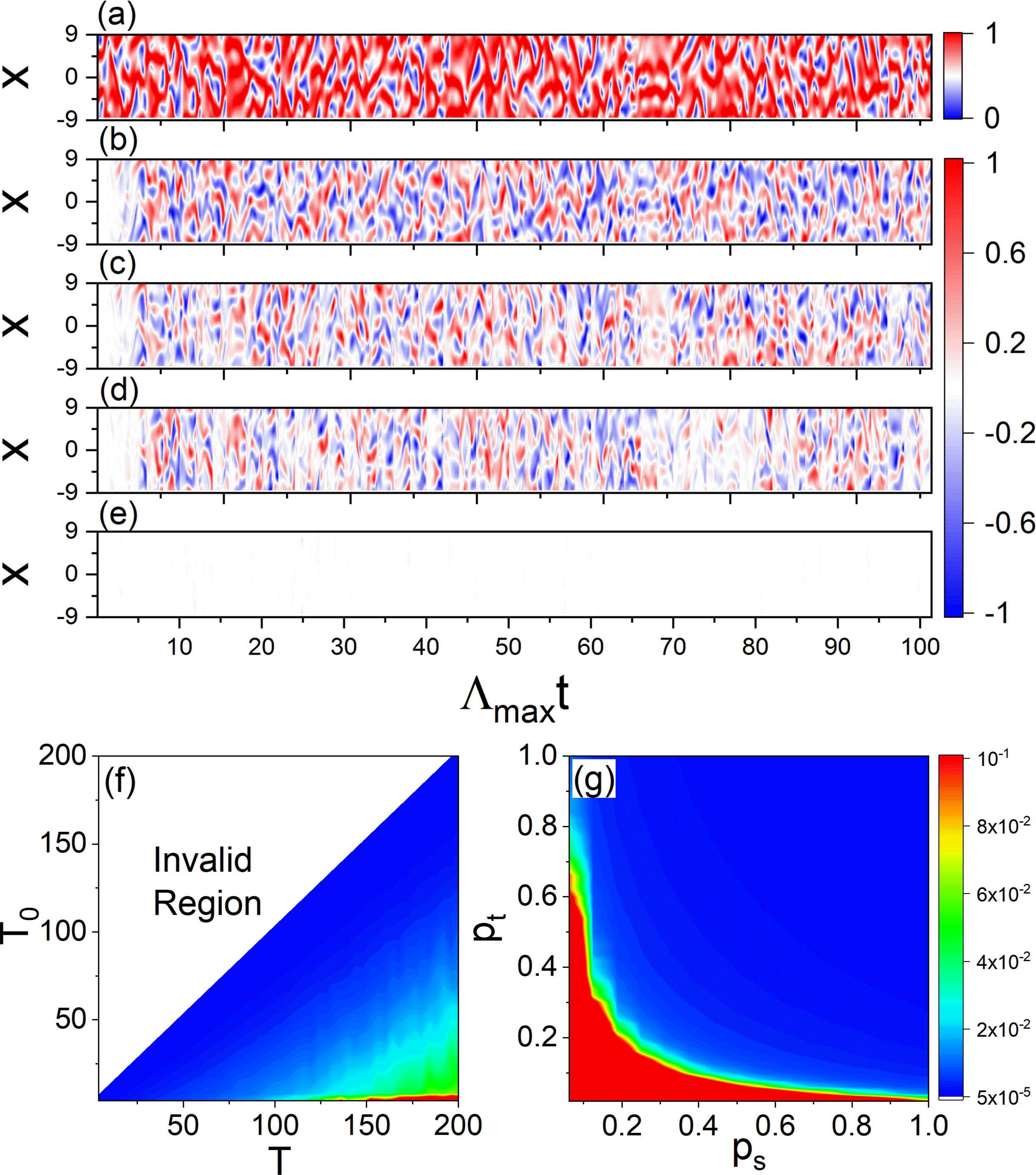}
\caption{ {\em Long time prediction of the state evolution of one-dimensional 
cGLE and }. The system size is $L=18$ and the number of measurement sites is 
$M=32$. (a) Spatiotemporal pattern of true state evolution. (b-e) The 
difference between the predicted and actual state evolution for $T=\infty$, 
$T=195$, $T=130$, and $T=65$, respectively. (f) For regular updating, 
prediction error in the parameter plane $(T_{0},T)$ for $M_c=32$ and $c=0.4$. 
(g) For random updating, prediction error in the parameter plane 
$(p_{t},p_{s})$ for $c=0.4$.}
\label{fig:CGLE}
\end{figure*}


We now demonstrate the predictive power of our reservoir computing scheme 
for the 1D cGLE in the regime of spatiotemporal chaos:     
$A_{t}=(1+i\alpha)A_{xx}+A-(1+i\beta)|A|^{2}A$, where $A(x,t)$ is a complex 
field in the interval $x\in(-L/2,L/2)$ with the periodic boundary condition,
$\alpha$ and $\beta$ are parameters. The cGLE is a general model for 
a variety of physical phenomena~\cite{AK:2002,CH:1993,Kbook:1984}. 
For $L=18$, $\alpha=2$ and $\beta = -2$, the 1D cGLE exhibits spatiotemporal 
chaos with the maximum Lyapunov exponent $\Lambda \approx 0.23$. We divide
the whole interval into $M = 32$ equally spaced points - the measurement sites.
For integration step $\Delta t=0.07$, one Lyapunov time corresponds to about
65 time steps. Figure~\ref{fig:CGLE}(a) shows the true spatiotemporal 
evolution pattern $|A(x,t)|$. Because the field $A(x,t)$ is complex, it is
necessary to use both the real [$A^r(x,t)$] and imaginary [$A^i(x,t)$] parts 
for training the reservoir system. The parameters of the reservoir system 
are $D_{in}=D_{out}=2M$, $D_{r}=9984$, $\sigma=1$, 
$d=3$, $\rho=0.1$, and $\eta=2\times10^{-5}$. Without any update, the error
between the predicted and true state evolution is shown in 
Fig.~\ref{fig:CGLE}(b), where the prediction horizon is about five Lyapunov
time. To introduce rare updates of the true state, we use the coupling scheme
$A'^{r(i)}_{rc}(x,t) = A^{r(i)}_{rc}(x,t) + 
0.4[A^{r(i)}_{act}(x,t)-A^{r(i)}_{rc}(x,t)]$ if there is actual data at 
the space-time point $(x,t)$, and otherwise 
$A'^{r(i)}_{rc}(x,t)=A^{r(i)}_{rc}(x,t)$, where 
$A^{r(i)}_{act}(x,t)$ is the real (imaginary) part of the true data and  
$A^{r(i)}_{rc}$ is the real (imaginary) part of predicted state. 
We demonstrate updating with only a single data point: $T_{0}=1$. 
Figures~\ref{fig:CGLE}(c-e) show the prediction error for $T=195$, 130, and
65, corresponding to three, two, and one Lyapunov time, respectively. Evidence 
of intermittent synchronization between the reservoir and the 
actual systems is shown in Figs.~\ref{fig:CGLE}(c) and \ref{fig:CGLE}(d).
As for the spatiotemporally chaotic KSE system, updating the reservoir system 
with a single data point every Lyapunov time [Fig.~\ref{fig:CGLE}(e)] leads to
arbitrarily long prediction horizon for the spatiotemporally chaotic state
of the 1D cGLE.

Figures~\ref{fig:CGLE}(f) and \ref{fig:CGLE}(g) show the time averaged 
prediction error $\delta e$ in the parameter planes $(T_{0},T)$ (regular 
updating scheme) and $(p_s,p_t)$ (random updating scheme), respectively, 
where the average is taken over approximately 100 Lyapunov time. For
regular updating, small prediction error can be achieved in most of the
parameter plane. For random updating, the error decreases with an increase 
in $p_t$ and/or $p_s$. 

The theoretical explanation for the observed long-term prediction is that,
after proper training the output vector of the reservoir system follows the 
state vector of the target system for a finite amount of time, indicating 
complete synchronization between the two systems. Without state updating, 
the synchronization state is slightly unstable, leading to a short 
prediction horizon. State updates, even applied rarely, represent a kind of 
perturbation that makes the synchronization state less unstable, prolonging 
the prediction horizon. When the frequency of the updates is such that there 
is one update within one Lyapunov time, the synchronization state becomes 
stable, giving rise to an arbitrarily long prediction horizon. This scenario 
has been verified using low-dimensional chaotic systems (Appendix A) through a 
synchronization stability analysis~\cite{HPC:1995,PC:1998}. We also note that, 
for a variety of low-dimensional chaotic systems including the classic 
R\"{o}ssler~\cite{Rossler:1976} and Lorenz~\cite{Lorenz:1963} oscillators, the 
Hindmarsh-Rose neuron~\cite{HR:1984}, and a chaotic food web~\cite{BHS:1999}, 
rare updates to a single state variable enables a properly trained reservoir 
system to predict {\em all} state variables for an arbitrarily long period 
of time (Appendices B-D). For example, for the chaotic food web, it is 
necessary only to supply sparse vegetation data for the reservoir system to 
correctly predict the abundances of the herbivores and predators, for as long 
as one wishes. 

To summarize, existing reservoir computing systems can predict the dynamical 
evolution of chaotic systems but only for a short period of time. We have 
articulated a scheme incorporating true state updates and demonstrated that 
rare updates of a subset of state variables can significantly prolong the 
prediction horizon. Of particular interest is the finding that, insofar as 
there is state update of a single data point within one Lyapunov time, the 
prediction horizon can be made arbitrarily long. The machine learning scheme 
proposed and studied here has the potential to extend significantly the 
application scope of reservoir computing in predicting complex dynamical 
systems. 

The novelty of our work is threefold: (1) we have introduced intermittent
data updates into machine learning, (2) we have demonstrated that 
long prediction time can be achieved, and (3) we have introduced the concept 
of temporal synchronization with on-off coupling to understand the working of 
the reservoir computing machine with state updating. Our work sheds new light 
in the working of reservoir computing in predicting the state evolution of 
chaotic systems. In particular, in the synchronization-based scenario, the 
predictability of reservoir computing is the result of its ability to 
synchronize with the target chaotic system for a finite amount of time.
At the start of the prediction phase, the reservoir system has the same
initial condition as the target system. Because of temporal synchronizability,
the reservoir system is able to follow the target system for sometime before
the synchronization error becomes significant. Without state updates, the
time it takes for the error to grow to a predefined threshold value determines
the prediction time. State updates, even being rare, reset the synchronization
error from time to time, insofar as a new update is provided before the error
exceeds the threshold.

In essence, the basic idea of our method is similar to that of data 
assimilation, where models and measurements (true state updates) are 
combined to generate accurate predictions with applications in, e.g.,
weather forecasting~\cite{PHKYO:2001,Ottetal:2004,SKGPHKOY:2005}.
In a recent study, data assimilation and machine learning have been combined 
to emulate the Lorenz 96 model from sparse and noisy 
observations~\cite{JAML:2019}. It is also noteworthy that, 
in predicting chaotic dynamical systems, an alternative 
machine-learning based framework is radial basis function networks - 
artificial neural networks employing radial basis functions as activation 
functions~\cite{BL:1988,PG:1990}. Given a set of inputs, such a network 
outputs a signal that is a linear combination of radial basis functions 
of the inputs, and the parameters of the artificial neurons are determined 
through training based on, e.g., the standard backpropagation scheme. The 
method has been demonstrated to be effective for low-dimensional chaotic 
systems such as the logistic map and the classic Lorenz chaotic 
oscillator~\cite{GAL:2006,CH:2013,RS:2016,ND:2018}. Whether the method can 
be superior to reservoir computing in predicting spatiotemporal chaotic 
systems is an open question. 

\section*{Acknowledgment}

The first two authors contributed equally to this work. This work is supported 
by the Office of Naval Research through Grant No.~N00014-16-1-2828.

\appendix

\section{Mechanism for reservoir computing to achieve arbitrarily long prediction horizon - a heuristic analysis based on synchronization}

\begin{figure}[ht!]
\centering
\includegraphics[width=\linewidth]{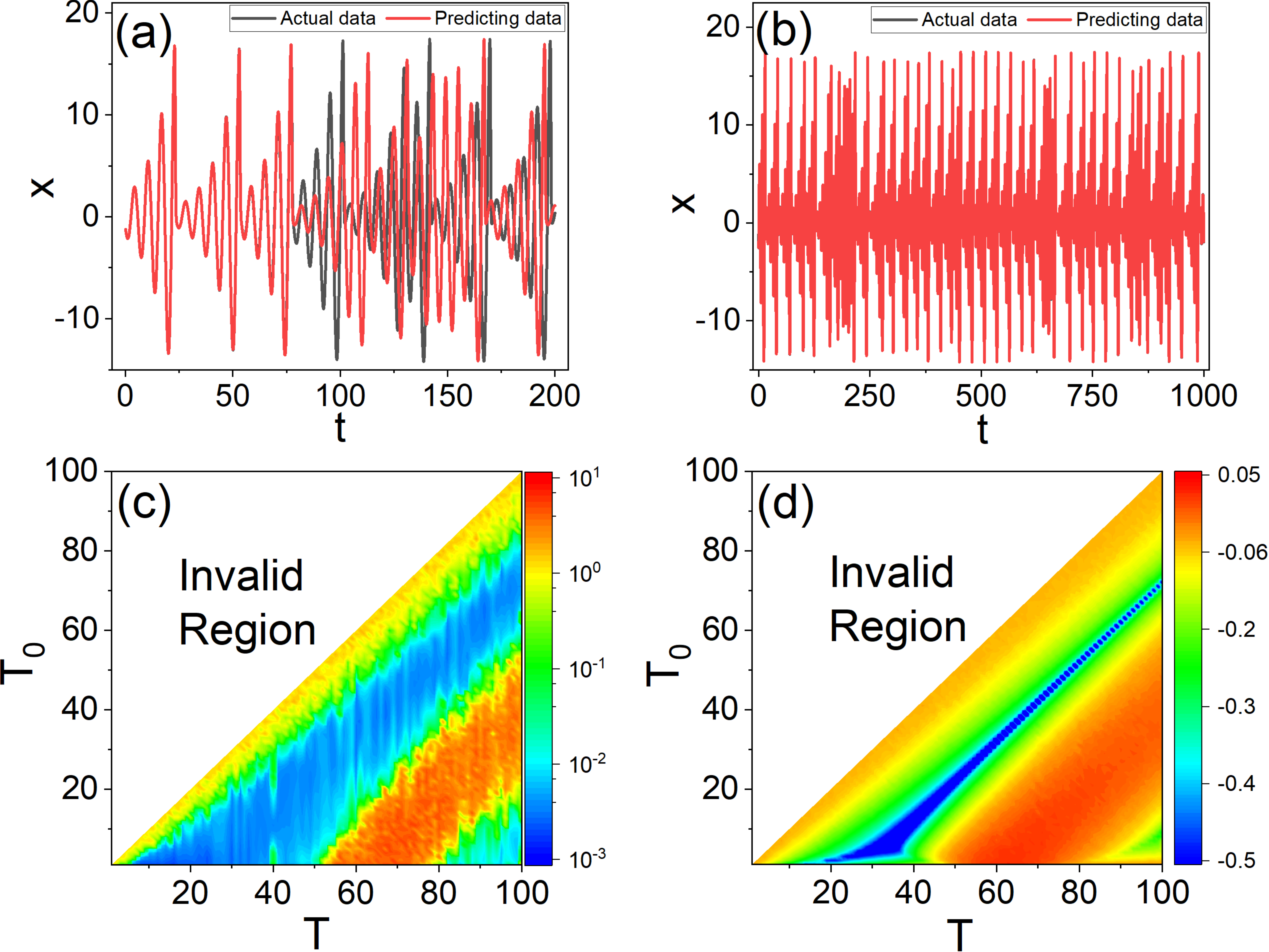}
\caption{ {\em Predicting state evolution of chaotic R\"{o}ssler oscillator 
and synchronization analysis}. (a) Result of prediction with conventional 
reservoir computing: true (black) and predicted (red) time series $x(t)$. The 
prediction horizon is approximately 15 average periods of oscillation as 
the true and predicted time series begin to diverge after this time. (b)
True and predicted time series with our proposed reservoir computing scheme
incorporating rare state updating. The two time series overlap completely 
and the prediction horizon is practically infinite. (c) Color-coded prediction
error in the parameter plane $(T_{0},T)$, where $T$ is the updating period
(in units of $\Delta t$, the time step between two successive iterations of 
the reservoir neural network) and $T_0 < T$ is the number of time steps 
during which there is true data input. (d) The maximum transverse Lyapunov 
exponent in $(T_{0},T)$ from synchronization analysis. There is a qualitative
correspondence between the small error [(c)] and negative Lyapunov exponent 
regions in the parameter plane.}
\label{fig:rossler}
\end{figure}

To understand the mechanism behind the realization of long prediction horizon
in our proposed reservoir computing scheme with state updating, we resort to 
synchronization analysis. To facilitate the analysis, we apply our scheme to
the chaotic R\"{o}ssler oscillator given by~\cite{Rossler:1976}:
\begin{align} \label{rossler}
&\dot{x}=-y-z \notag, \\
&\dot{y}=x+0.2y , \\
&\dot{z}=0.2+(x-9.0)z \notag.
\end{align}
We obtain the three time series [$x(t)$, $y(t)$, $z(t)$] at the step size 
$\Delta t=0.05$. The values of the parameters of the reservoir computing 
system are  $D_{in}=D_{out}=3$, $D_{r}=600$, $\sigma=0.15$, $d'=0.2$, 
$\rho=0.2$, and $\eta=1\times 10^{-7}$, where $d'$ is the link density of the 
complex neural network. The elements of the $D_{r}\times D_{in}$ input matrix 
$\bm{W}_{in}$ are generated from the uniform distribution $[-\sigma,\sigma]$. 
The non-zero elements of the $D_r\times D_r$ matrix $\bm{A}$ are generated 
from the uniform distribution $[-1,1]$. Figure~\ref{fig:rossler}(a) shows the 
prediction result with the conventional scheme, where the prediction horizon 
is about $t \approx 75$ (corresponding to approximately 15 average cycles of 
oscillation). Figure~\ref{fig:rossler}(b) shows that, with our proposed 
scheme, the prediction horizon is practically infinite, where an update of the 
actual data of {\em a single dynamical variable}, $y_{act}$, is coupled into 
the system ($T_0=1$ and $T=50$) once every 50 time steps. More specifically, 
in the iterative process of reservoir computing system, the input data $y_{rc}$
is replaced by $y'_{rc}=y_{rc}+c(y_{act}-y_{rc})$ once every 50 time steps 
for $c=0.8$. Note that, in Fig.~\ref{fig:rossler}(b), only the true and 
predicted $x$ time series are shown, but time series from the other two 
dynamical variables give essentially the same prediction result. 
Figure~\ref{fig:rossler}(c) shows the color-coded average prediction error 
$\delta e$ in the parameter plane $(T_{0},T)$, where there are multiple 
parameter regions in which the error is small. The patterns in 
Fig.~\ref{fig:rossler}(c) are reminiscent of the phenomenon of ragged 
synchronization in coupled chaotic oscillator systems~\cite{SPK:2007,PJSK:2008},
suggesting the use of synchronization theory to understand the working 
mechanism of our articulated reservoir computing machine.

A well trained reservoir computing system can be viewed as a high-dimensional
replica of the target system. A previous calculation of the Lyapunov exponents
of the reservoir dynamical network revealed that the first few exponents are
indeed approximate values of the exponents of the true system, and the 
vast set of remaining exponents have large negative values~\cite{PLHGO:2017}. 
This is anticipated as the large negative exponents are necessary to reduce 
the exceedingly high dimension of the reservoir network to the low-dimensional 
target system through a strong compression of the dynamics along vast majority 
of orthogonal directions in the phase space. For the chaotic R\"{o}ssler 
system, the considerations suggest that the dynamics of the reservoir 
computing system be approximately described by
\begin{align}\label{RC:coupling}
&\dot{x}_{rc}=-y_{rc}-z_{rc}, \\ \nonumber
&\dot{y}_{rc}=x_{rc}+0.2y_{rc}+\varepsilon(t)(y_{act}-y_{rc}), \\
&\dot{z}_{rc}=0.2+(x_{rc}-9)z_{rc}, \notag
\end{align}
where, in the coupling term $\varepsilon(t)(y_{act}-y_{rc})$, $\varepsilon(t)$ 
specifies the on-off nature of the coupling: $\varepsilon(t)=\varepsilon$ if 
$nT\Delta t<t<T_{0}\Delta t+nT\Delta t$ ($n=0,1,2,\ldots$) and 
$\varepsilon(t)=0$ otherwise. There is a linear relation between 
$\varepsilon$ and $c$: $c=\varepsilon\Delta t$. The data of 
$y_{act}$ are generated from the target system Eq.~(\ref{rossler}). 
In our scheme, predictability implies synchronization between the reduced 
reservoir computing system Eq.~(\ref{RC:coupling}) and the true system 
Eq.~(\ref{rossler}) with only on-off coupling, where the ``on'' phase is 
typically significantly shorter than the ``off'' phase.

We use stability analysis to quantify synchronization. Let 
$\delta x=x_{rc}-x_{act}$, $\delta y=y_{rc}-y_{act}$, and 
$\delta z=z_{rc}-z_{act}$ be the infinitesimal perturbations transverse to 
the synchronization manifold. The variational equations can be obtained by 
linearizing Eq.~(\ref{RC:coupling}) with respect to the true state of the 
target system as described by Eq.~(\ref{rossler}):
\begin{align} \label{Jacobian}
&\delta\dot{x}=-\delta y-\delta z \notag, \\
&\delta\dot{y}=\delta x+0.2\delta y -\varepsilon(t)\delta y, \\
&\delta\dot{z}=z\delta x +(x-9.0)\delta z\notag.
\end{align}
Combining Eqs.~(\ref{rossler}) and (\ref{Jacobian}), we calculate the maximum 
transverse Lyapunov exponent $\Lambda$. Stable synchronization requires 
$\Lambda < 0$. Figure~\ref{fig:rossler}(d) shows the color-coded value of 
$\Lambda$ in the parameter plane $(T_{0},T)$, which exhibits typical features
of ragged synchronization~\cite{SPK:2007,PJSK:2008}. Comparing 
Fig.~\ref{fig:rossler}(c) with Fig.~\ref{fig:rossler}(d), we observe a 
striking degree of similarity, indicating synchronization between the reservoir
computing and the target systems subject to on-off coupling as the dynamical 
mechanism responsible for realizing the long prediction horizon with rare data 
updating.

\begin{figure}[ht!]
\centering
\includegraphics[width=\linewidth]{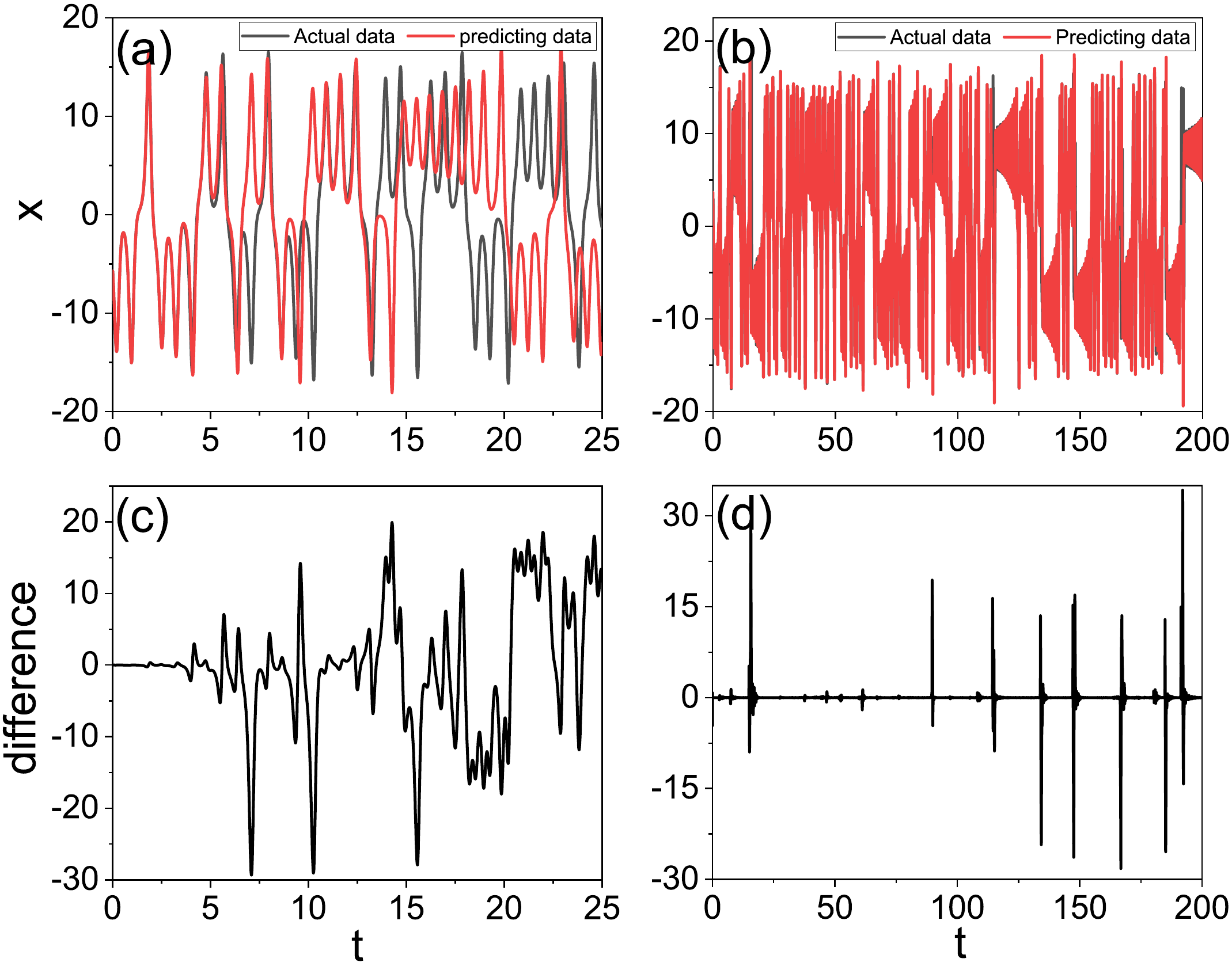}
\caption{ {\em Predicting the chaotic Lorenz system with rare state updating}. 
(a) Prediction with the conventional reservoir computing scheme without 
state updating, where the prediction horizon is about ten oscillations.
(b) Practically infinite prediction horizon achieved with sparse input of 
actual data for $T_{0}=1$, $T=40$, and $c=0.8$. (c,d) The difference between 
the prediction and actual time series corresponding to the results in 
(a,b), respectively.}
\label{fig:lorenz}
\end{figure}

\section{Predicting chaotic Lorenz system}

We demonstrate that a practically infinite prediction horizon can be achieved
for the classic chaotic Lorenz system~\cite{Lorenz:1963} with rare state 
updating. The equations of the system are 
\begin{align} \label{lorenz}
&\dot{x}=10(y-x) \notag, \\
&\dot{y}=x(28-z)-y , \\
&\dot{z}=xy-8/3z\notag.
\end{align}
We obtain the time series $x(t)$, $y(t)$, and $z(t)$ with integration
step size $\Delta t=0.01$. The parameter setting of the reservoir computing
system is $D_{in}=D_{out}=3$, $D_{r}=600$, $\sigma=0.1$, $d'=0.3$, $\rho=1.2$, 
and $\eta=1\times 10^{-5}$. Figure~\ref{fig:lorenz}(a) shows the prediction
result with the conventional scheme without any state updating, where the 
prediction horizon is $t\approx 5$ (corresponding to approximately ten average
oscillations). Figure~\ref{fig:lorenz}(b) shows that, with rare state updating
($T_{0}=1$ and $T=40$) of one of the dynamical variables, mathematically 
represented as replacement of $y_{rc}$ by $y'_{rc}=y_{rc}+c(y_{act}-y_{rc})$
once every 40 time steps, practically an arbitrarily long prediction horizon 
can be achieved. Figures~\ref{fig:lorenz}(c) and \ref{fig:lorenz}(d) show the 
difference between the predicted and true time series from the conventional 
and our proposed reservoir computing schemes, respectively. It can be seen
that, with rare state updating, the prediction error is essentially zero 
for the time interval displayed, with relatively large errors occurring only 
at about a few dozen time steps (out of $2\times 10^4$ time steps).  

\begin{figure}[ht!]
\centering
\includegraphics[width=\linewidth]{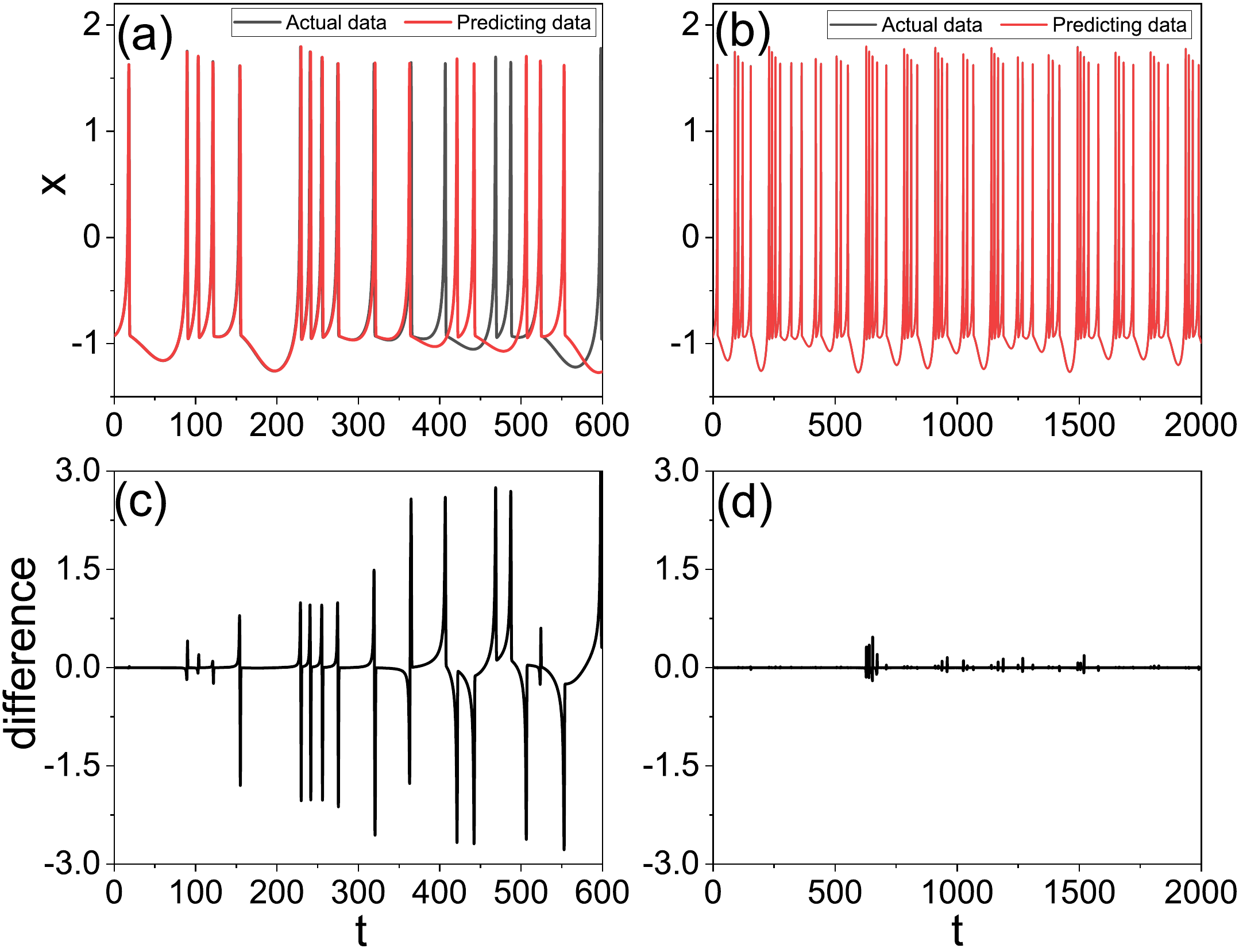}
\caption{{\em Predicting the state evolution of a chaotic Hindmarsh-Rose 
neuron}. (a) Prediction without state updating. (b) Arbitrarily long prediction
horizon with rare state updating for $T_{0}=1$, $T=300$, and $c=0.9$. (c,d) 
The difference between the predicted and true state evolution for the cases
in (a,b), respectively.} 
\label{fig:hr}
\end{figure}

\begin{figure}[ht!]
\centering
\includegraphics[width=\linewidth]{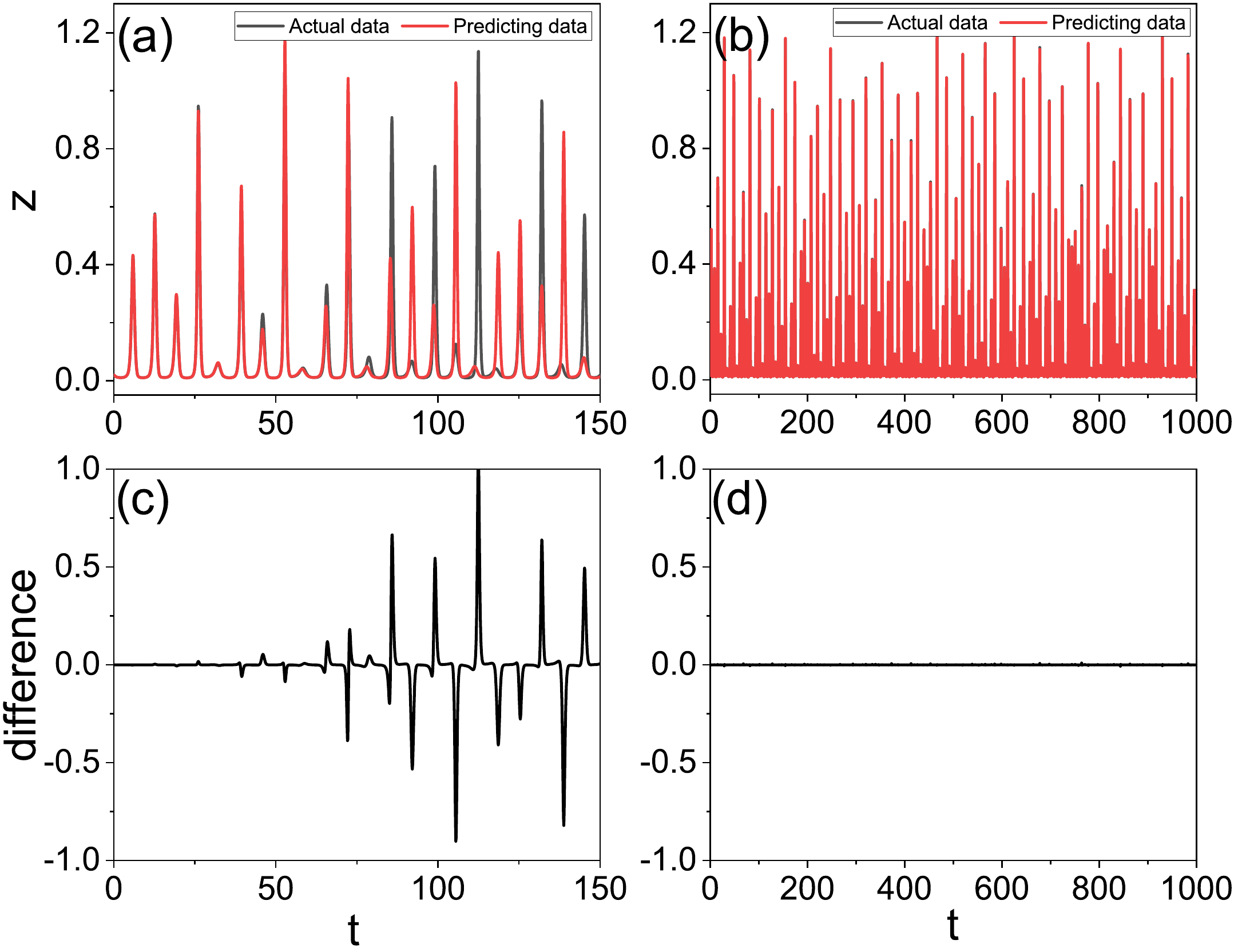}
\caption{{\em Predicting the state evolution of the populations of a chaotic 
food web}. (a) Predicted and actual state evolution with the conventional 
reservoir computing scheme without any state updating. (b) Prediction result
with rate state updating: once every 50 time steps ($T_{0}=1$ and $T=50$). 
The value of the coupling parameter is $c=1$. (c,d) Evolution of the 
prediction error corresponding to the cases in (a,b), respectively. With the 
rare state updating, the prediction error is practically zero in the entire 
time interval of $10^4$ time steps tested.}
\label{fig:fw}
\end{figure}

\section{Predicting chaotic Hindmarsh-Rose neuron dynamics}

We test our reservoir computing scheme with rare state updating for the 
chaotic Hindmarsh-Rose neuron model~\cite{HR:1984}: 
\begin{align} \label{hr}
&\dot{x}=y+3x^{2}-x^{3}-z+3.2, \notag \\
&\dot{y}=1-5x^{2}-y, \\
&\dot{z}=-0.006z+0.024(x+1.6), \notag
\end{align}
where $x$ is the membrane potential, $y$ and $z$ are the transport rates of 
the fast and slow channels, respectively. The integration time step is 
$\Delta t=0.1$. The three time series $x(t)$, $y(t)$, and $z(t)$ are used 
to train the reservoir computing system with parameters $D_{in}=D_{out}=3$, 
$D_{r}=600$, $\sigma=0.6$, $d'=0.2$, $\rho=0.3$, and $\eta=1\times 10^{-7}$. 
Figure~\ref{fig:hr}(a) shows the result with the conventional reservoir 
computing scheme without any state updating, where the prediction horizon is 
$t\approx 300$, in which there are about a dozen spiking events. With rare
state updating (once every 100 time steps: $T_{0}=1$ and $T=100$) of two 
state variables ($x_{act}$ and $y_{act}$), the prediction horizon is 
practically infinite. Mathematically, the updating scheme can be described
as a single replacement every 100 time steps of $x_{rc}$ and $y_{rc}$ by 
$x'_{rc}=x_{rc}+c(x_{act}-x_{rc})$ and $y'_{rc}=y_{rc}+c(y_{act}-y_{rc})$
respectively, for $c = 0.9$. The prediction errors corresponding to 
Figs.~\ref{fig:hr}(a) and \ref{fig:hr}(b) are shown in Figs.~\ref{fig:hr}(c)
and \ref{fig:hr}(d), respectively. 

\section{Predicting population evolution in a chaotic food web}

We demonstrate an arbitrarily long prediction horizon for the
following chaotic food web system~\cite{BHS:1999}:
\begin{align}\label{fw}
&\dot{x}=x-0.2\frac{xy}{1+0.05x}, \notag \\
&\dot{y}=-y+0.2\frac{xy}{1+0.05x}-yz, \\
&\dot{z}=-10(z-0.006)+yz, \notag
\end{align}
where $x$, $y$ and $z$ represent vegetation, herbivores and predators, and 
the evolution displays uniform phase evolution but with chaotic amplitude
modulation. The integration step size is $\Delta t=0.1$. The parameter values
of the reservoir computing system are $D_{in}=D_{out}=3$, $D_{r}=600$, 
$\sigma=0.15$, $d'=0.2$, $\rho=0.2$, and $\eta=1\times 10^{-7}$. 
Figure~\ref{fig:fw}(a) shows the prediction result from the conventional
scheme without any state updating, where the prediction horizon is 
$t\approx 50$ (containing seven or eight bursts in the predator population).
In the food web system, vegetation data are relatively easy to be collected,
so we use $x_{act}$ to perform rare state updating. Figure~\ref{fig:fw}(b)
shows the predicted and actual predator time series for $T_{0}=1$ and $T=50$,
i.e., we replace $x_{rc}$ by $x'_{rc}=x_{rc}+c(x_{act}-x_{rc})$ once every
50 time steps. Visually the two types of time series cannot be distinguished.
Figures~\ref{fig:fw}(c) and \ref{fig:fw}(d) show the corresponding evolution
of the prediction error for Figs.~\ref{fig:fw}(a) and \ref{fig:fw}(b),
respectively. We see that, with rare state updating, the prediction error
is exceedingly small in the long time interval ($10^4$ time steps) tested,
indicating that a practically infinite prediction horizon has been achieved.


\begin{thebibliography}{40}%
\makeatletter
\providecommand \@ifxundefined [1]{%
 \@ifx{#1\undefined}
}%
\providecommand \@ifnum [1]{%
 \ifnum #1\expandafter \@firstoftwo
 \else \expandafter \@secondoftwo
 \fi
}%
\providecommand \@ifx [1]{%
 \ifx #1\expandafter \@firstoftwo
 \else \expandafter \@secondoftwo
 \fi
}%
\providecommand \natexlab [1]{#1}%
\providecommand \enquote  [1]{``#1''}%
\providecommand \bibnamefont  [1]{#1}%
\providecommand \bibfnamefont [1]{#1}%
\providecommand \citenamefont [1]{#1}%
\providecommand \href@noop [0]{\@secondoftwo}%
\providecommand \href [0]{\begingroup \@sanitize@url \@href}%
\providecommand \@href[1]{\@@startlink{#1}\@@href}%
\providecommand \@@href[1]{\endgroup#1\@@endlink}%
\providecommand \@sanitize@url [0]{\catcode `\\12\catcode `\$12\catcode
  `\&12\catcode `\#12\catcode `\^12\catcode `\_12\catcode `\%12\relax}%
\providecommand \@@startlink[1]{}%
\providecommand \@@endlink[0]{}%
\providecommand \url  [0]{\begingroup\@sanitize@url \@url }%
\providecommand \@url [1]{\endgroup\@href {#1}{\urlprefix }}%
\providecommand \urlprefix  [0]{URL }%
\providecommand \Eprint [0]{\href }%
\providecommand \doibase [0]{http://dx.doi.org/}%
\providecommand \selectlanguage [0]{\@gobble}%
\providecommand \bibinfo  [0]{\@secondoftwo}%
\providecommand \bibfield  [0]{\@secondoftwo}%
\providecommand \translation [1]{[#1]}%
\providecommand \BibitemOpen [0]{}%
\providecommand \bibitemStop [0]{}%
\providecommand \bibitemNoStop [0]{.\EOS\space}%
\providecommand \EOS [0]{\spacefactor3000\relax}%
\providecommand \BibitemShut  [1]{\csname bibitem#1\endcsname}%
\let\auto@bib@innerbib\@empty
\bibitem [{\citenamefont {Haynes}\ \emph {et~al.}(2015)\citenamefont {Haynes},
  \citenamefont {Soriano}, \citenamefont {Rosin}, \citenamefont {Fischer},\
  and\ \citenamefont {Gauthier}}]{HSRFG:2015}%
  \BibitemOpen
  \bibfield  {author} {\bibinfo {author} {\bibfnamefont {N.~D.}\ \bibnamefont
  {Haynes}}, \bibinfo {author} {\bibfnamefont {M.~C.}\ \bibnamefont {Soriano}},
  \bibinfo {author} {\bibfnamefont {D.~P.}\ \bibnamefont {Rosin}}, \bibinfo
  {author} {\bibfnamefont {I.}~\bibnamefont {Fischer}}, \ and\ \bibinfo
  {author} {\bibfnamefont {D.~J.}\ \bibnamefont {Gauthier}},\ }\bibfield
  {title} {\enquote {\bibinfo {title} {Reservoir computing with a single
  time-delay autonomous {Boolean} node},}\ }\href {\doibase
  10.1103/PhysRevE.91.020801} {\bibfield  {journal} {\bibinfo  {journal} {Phys.
  Rev. E}\ }\textbf {\bibinfo {volume} {91}},\ \bibinfo {pages} {020801}
  (\bibinfo {year} {2015})}\BibitemShut {NoStop}%
\bibitem [{\citenamefont {Larger}\ \emph {et~al.}(2017)\citenamefont {Larger},
  \citenamefont {Bayl\'on-Fuentes}, \citenamefont {Martinenghi}, \citenamefont
  {Udaltsov}, \citenamefont {Chembo},\ and\ \citenamefont
  {Jacquot}}]{LBMUCJ:2017}%
  \BibitemOpen
  \bibfield  {author} {\bibinfo {author} {\bibfnamefont {L.}~\bibnamefont
  {Larger}}, \bibinfo {author} {\bibfnamefont {A.}~\bibnamefont
  {Bayl\'on-Fuentes}}, \bibinfo {author} {\bibfnamefont {R.}~\bibnamefont
  {Martinenghi}}, \bibinfo {author} {\bibfnamefont {V.~S.}\ \bibnamefont
  {Udaltsov}}, \bibinfo {author} {\bibfnamefont {Y.~K.}\ \bibnamefont
  {Chembo}}, \ and\ \bibinfo {author} {\bibfnamefont {M.}~\bibnamefont
  {Jacquot}},\ }\bibfield  {title} {\enquote {\bibinfo {title} {High-speed
  photonic reservoir computing using a time-delay-based architecture: Million
  words per second classification},}\ }\href {\doibase
  10.1103/PhysRevX.7.011015} {\bibfield  {journal} {\bibinfo  {journal} {Phys.
  Rev. X}\ }\textbf {\bibinfo {volume} {7}},\ \bibinfo {pages} {011015}
  (\bibinfo {year} {2017})}\BibitemShut {NoStop}%
\bibitem [{\citenamefont {Pathak}\ \emph {et~al.}(2017)\citenamefont {Pathak},
  \citenamefont {Lu}, \citenamefont {Hunt}, \citenamefont {Girvan},\ and\
  \citenamefont {Ott}}]{PLHGO:2017}%
  \BibitemOpen
  \bibfield  {author} {\bibinfo {author} {\bibfnamefont {J.}~\bibnamefont
  {Pathak}}, \bibinfo {author} {\bibfnamefont {Z.}~\bibnamefont {Lu}}, \bibinfo
  {author} {\bibfnamefont {B.}~\bibnamefont {Hunt}}, \bibinfo {author}
  {\bibfnamefont {M.}~\bibnamefont {Girvan}}, \ and\ \bibinfo {author}
  {\bibfnamefont {E.}~\bibnamefont {Ott}},\ }\bibfield  {title} {\enquote
  {\bibinfo {title} {Using machine learning to replicate chaotic attractors and
  calculate {Lyapunov} exponents from data},}\ }\href@noop {} {\bibfield
  {journal} {\bibinfo  {journal} {Chaos}\ }\textbf {\bibinfo {volume} {27}},\
  \bibinfo {pages} {121102} (\bibinfo {year} {2017})}\BibitemShut {NoStop}%
\bibitem [{\citenamefont {Lu}\ \emph {et~al.}(2017)\citenamefont {Lu},
  \citenamefont {Pathak}, \citenamefont {Hunt}, \citenamefont {Girvan},
  \citenamefont {Brockett},\ and\ \citenamefont {Ott}}]{LPHGBO:2017}%
  \BibitemOpen
  \bibfield  {author} {\bibinfo {author} {\bibfnamefont {Z.}~\bibnamefont
  {Lu}}, \bibinfo {author} {\bibfnamefont {J.}~\bibnamefont {Pathak}}, \bibinfo
  {author} {\bibfnamefont {B.}~\bibnamefont {Hunt}}, \bibinfo {author}
  {\bibfnamefont {M.}~\bibnamefont {Girvan}}, \bibinfo {author} {\bibfnamefont
  {R.}~\bibnamefont {Brockett}}, \ and\ \bibinfo {author} {\bibfnamefont
  {E.}~\bibnamefont {Ott}},\ }\bibfield  {title} {\enquote {\bibinfo {title}
  {Reservoir observers: Model-free inference of unmeasured variables in chaotic
  systems},}\ }\href@noop {} {\bibfield  {journal} {\bibinfo  {journal}
  {Chaos}\ }\textbf {\bibinfo {volume} {27}},\ \bibinfo {pages} {041102}
  (\bibinfo {year} {2017})}\BibitemShut {NoStop}%
\bibitem [{\citenamefont {Duriez}\ \emph {et~al.}(2017)\citenamefont {Duriez},
  \citenamefont {Brunton},\ and\ \citenamefont {Noack}}]{DBN:book}%
  \BibitemOpen
  \bibfield  {author} {\bibinfo {author} {\bibfnamefont {T.}~\bibnamefont
  {Duriez}}, \bibinfo {author} {\bibfnamefont {S.~L.}\ \bibnamefont {Brunton}},
  \ and\ \bibinfo {author} {\bibfnamefont {B.~R.}\ \bibnamefont {Noack}},\
  }\href@noop {} {\emph {\bibinfo {title} {Machine Learning Control-Taming
  Nonlinear Dynamics and Turbulence}}}\ (\bibinfo  {publisher} {Springer},\
  \bibinfo {year} {2017})\BibitemShut {NoStop}%
\bibitem [{\citenamefont {Pathak}\ \emph
  {et~al.}(2018{\natexlab{a}})\citenamefont {Pathak}, \citenamefont {Wilner},
  \citenamefont {Fussell}, \citenamefont {Chandra}, \citenamefont {Hunt},
  \citenamefont {Girvan}, \citenamefont {Lu},\ and\ \citenamefont
  {Ott}}]{PWFCHGO:2018}%
  \BibitemOpen
  \bibfield  {author} {\bibinfo {author} {\bibfnamefont {J.}~\bibnamefont
  {Pathak}}, \bibinfo {author} {\bibfnamefont {A.}~\bibnamefont {Wilner}},
  \bibinfo {author} {\bibfnamefont {R.}~\bibnamefont {Fussell}}, \bibinfo
  {author} {\bibfnamefont {S.}~\bibnamefont {Chandra}}, \bibinfo {author}
  {\bibfnamefont {B.}~\bibnamefont {Hunt}}, \bibinfo {author} {\bibfnamefont
  {M.}~\bibnamefont {Girvan}}, \bibinfo {author} {\bibfnamefont
  {Z.}~\bibnamefont {Lu}}, \ and\ \bibinfo {author} {\bibfnamefont
  {E.}~\bibnamefont {Ott}},\ }\bibfield  {title} {\enquote {\bibinfo {title}
  {Hybrid forecasting of chaotic processes: Using machine learning in
  conjunction with a knowledge-based model},}\ }\href@noop {} {\bibfield
  {journal} {\bibinfo  {journal} {Chaos}\ }\textbf {\bibinfo {volume} {28}},\
  \bibinfo {pages} {041101} (\bibinfo {year} {2018}{\natexlab{a}})}\BibitemShut
  {NoStop}%
\bibitem [{\citenamefont {Pathak}\ \emph
  {et~al.}(2018{\natexlab{b}})\citenamefont {Pathak}, \citenamefont {Hunt},
  \citenamefont {Girvan}, \citenamefont {Lu},\ and\ \citenamefont
  {Ott}}]{PHGLO:2018}%
  \BibitemOpen
  \bibfield  {author} {\bibinfo {author} {\bibfnamefont {J.}~\bibnamefont
  {Pathak}}, \bibinfo {author} {\bibfnamefont {B.}~\bibnamefont {Hunt}},
  \bibinfo {author} {\bibfnamefont {M.}~\bibnamefont {Girvan}}, \bibinfo
  {author} {\bibfnamefont {Z.}~\bibnamefont {Lu}}, \ and\ \bibinfo {author}
  {\bibfnamefont {E.}~\bibnamefont {Ott}},\ }\bibfield  {title} {\enquote
  {\bibinfo {title} {Model-free prediction of large spatiotemporally chaotic
  systems from data: A reservoir computing approach},}\ }\href {\doibase
  10.1103/PhysRevLett.120.024102} {\bibfield  {journal} {\bibinfo  {journal}
  {Phys. Rev. Lett.}\ }\textbf {\bibinfo {volume} {120}},\ \bibinfo {pages}
  {024102} (\bibinfo {year} {2018}{\natexlab{b}})}\BibitemShut {NoStop}%
\bibitem [{\citenamefont {Carroll}(2018)}]{Carroll:2018}%
  \BibitemOpen
  \bibfield  {author} {\bibinfo {author} {\bibfnamefont {T.~L.}\ \bibnamefont
  {Carroll}},\ }\bibfield  {title} {\enquote {\bibinfo {title} {Using reservoir
  computers to distinguish chaotic signals},}\ }\href {\doibase
  10.1103/PhysRevE.98.052209} {\bibfield  {journal} {\bibinfo  {journal} {Phys.
  Rev. E}\ }\textbf {\bibinfo {volume} {98}},\ \bibinfo {pages} {052209}
  (\bibinfo {year} {2018})}\BibitemShut {NoStop}%
\bibitem [{\citenamefont {Nakai}\ and\ \citenamefont {Saiki}(2018)}]{NS:2018}%
  \BibitemOpen
  \bibfield  {author} {\bibinfo {author} {\bibfnamefont {K.}~\bibnamefont
  {Nakai}}\ and\ \bibinfo {author} {\bibfnamefont {Y.}~\bibnamefont {Saiki}},\
  }\bibfield  {title} {\enquote {\bibinfo {title} {Machine-learning inference
  of fluid variables from data using reservoir computing},}\ }\href {\doibase
  10.1103/PhysRevE.98.023111} {\bibfield  {journal} {\bibinfo  {journal} {Phys.
  Rev. E}\ }\textbf {\bibinfo {volume} {98}},\ \bibinfo {pages} {023111}
  (\bibinfo {year} {2018})}\BibitemShut {NoStop}%
\bibitem [{\citenamefont {Roland}\ and\ \citenamefont
  {Parlitz}(2018)}]{ZP:2018}%
  \BibitemOpen
  \bibfield  {author} {\bibinfo {author} {\bibfnamefont {Z.~S.}\ \bibnamefont
  {Roland}}\ and\ \bibinfo {author} {\bibfnamefont {U.}~\bibnamefont
  {Parlitz}},\ }\bibfield  {title} {\enquote {\bibinfo {title} {Observing
  spatio-temporal dynamics of excitable media using reservoir computing},}\
  }\href@noop {} {\bibfield  {journal} {\bibinfo  {journal} {Chaos}\ }\textbf
  {\bibinfo {volume} {28}},\ \bibinfo {pages} {043118} (\bibinfo {year}
  {2018})}\BibitemShut {NoStop}%
\bibitem [{\citenamefont {Weng}\ \emph {et~al.}(2019)\citenamefont {Weng},
  \citenamefont {Yang}, \citenamefont {Gu}, \citenamefont {Zhang},\ and\
  \citenamefont {Small}}]{WYGZS:2019}%
  \BibitemOpen
  \bibfield  {author} {\bibinfo {author} {\bibfnamefont {T.}~\bibnamefont
  {Weng}}, \bibinfo {author} {\bibfnamefont {H.}~\bibnamefont {Yang}}, \bibinfo
  {author} {\bibfnamefont {C.}~\bibnamefont {Gu}}, \bibinfo {author}
  {\bibfnamefont {J.}~\bibnamefont {Zhang}}, \ and\ \bibinfo {author}
  {\bibfnamefont {M.}~\bibnamefont {Small}},\ }\bibfield  {title} {\enquote
  {\bibinfo {title} {Synchronization of chaotic systems and their
  machine-learning models},}\ }\href {\doibase 10.1103/PhysRevE.99.042203}
  {\bibfield  {journal} {\bibinfo  {journal} {Phys. Rev. E}\ }\textbf {\bibinfo
  {volume} {99}},\ \bibinfo {pages} {042203} (\bibinfo {year}
  {2019})}\BibitemShut {NoStop}%
\bibitem [{\citenamefont {Jiang}\ and\ \citenamefont {Lai}(2019)}]{JL:2019}%
  \BibitemOpen
  \bibfield  {author} {\bibinfo {author} {\bibfnamefont {J.}~\bibnamefont
  {Jiang}}\ and\ \bibinfo {author} {\bibfnamefont {Y.-C.}\ \bibnamefont
  {Lai}},\ }\bibfield  {title} {\enquote {\bibinfo {title} {Model-free
  prediction of spatiotemporal dynamical systems with recurrent neural
  networks: Role of network spectral radius},}\ }\href {\doibase
  10.1103/PhysRevResearch.1.033056} {\bibfield  {journal} {\bibinfo  {journal}
  {Phys. Rev. Research}\ }\textbf {\bibinfo {volume} {1}},\ \bibinfo {pages}
  {033056} (\bibinfo {year} {2019})}\BibitemShut {NoStop}%
\bibitem [{\citenamefont {Jaeger}(2001)}]{Jaeger:2001}%
  \BibitemOpen
  \bibfield  {author} {\bibinfo {author} {\bibfnamefont {H.}~\bibnamefont
  {Jaeger}},\ }\bibfield  {title} {\enquote {\bibinfo {title} {The “echo
  state” approach to analysing and training recurrent neural networks-with an
  erratum note},}\ }\href@noop {} {\bibfield  {journal} {\bibinfo  {journal}
  {Bonn, Germany: German National Research Center for Information Technology
  GMD Technical Report}\ }\textbf {\bibinfo {volume} {148}},\ \bibinfo {pages}
  {13} (\bibinfo {year} {2001})}\BibitemShut {NoStop}%
\bibitem [{\citenamefont {Mass}\ \emph {et~al.}(2002)\citenamefont {Mass},
  \citenamefont {Nachtschlaeger},\ and\ \citenamefont {Markram}}]{MNM:2002}%
  \BibitemOpen
  \bibfield  {author} {\bibinfo {author} {\bibfnamefont {W.}~\bibnamefont
  {Mass}}, \bibinfo {author} {\bibfnamefont {T.}~\bibnamefont
  {Nachtschlaeger}}, \ and\ \bibinfo {author} {\bibfnamefont {H.}~\bibnamefont
  {Markram}},\ }\bibfield  {title} {\enquote {\bibinfo {title} {Real-time
  computing without stable states: A new framework for neural computation based
  on perturbations},}\ }\href@noop {} {\bibfield  {journal} {\bibinfo
  {journal} {Neur. Comp.}\ }\textbf {\bibinfo {volume} {14}},\ \bibinfo {pages}
  {2531} (\bibinfo {year} {2002})}\BibitemShut {NoStop}%
\bibitem [{\citenamefont {Jaeger}\ and\ \citenamefont {Haas}(2004)}]{JH:2004}%
  \BibitemOpen
  \bibfield  {author} {\bibinfo {author} {\bibfnamefont {H.}~\bibnamefont
  {Jaeger}}\ and\ \bibinfo {author} {\bibfnamefont {H.}~\bibnamefont {Haas}},\
  }\bibfield  {title} {\enquote {\bibinfo {title} {Harnessing nonlinearity:
  Predicting chaotic systems and saving energy in wireless communication},}\
  }\href {\doibase 10.1126/science.1091277} {\bibfield  {journal} {\bibinfo
  {journal} {Science}\ }\textbf {\bibinfo {volume} {304}},\ \bibinfo {pages}
  {78} (\bibinfo {year} {2004})}\BibitemShut {NoStop}%
\bibitem [{\citenamefont {Manjunath}\ and\ \citenamefont
  {Jaeger}(2013)}]{MJ:2013}%
  \BibitemOpen
  \bibfield  {author} {\bibinfo {author} {\bibfnamefont {G.}~\bibnamefont
  {Manjunath}}\ and\ \bibinfo {author} {\bibfnamefont {H.}~\bibnamefont
  {Jaeger}},\ }\bibfield  {title} {\enquote {\bibinfo {title} {Echo state
  property linked to an input: Exploring a fundamental characteristic of
  recurrent neural networks},}\ }\href@noop {} {\bibfield  {journal} {\bibinfo
  {journal} {Neur. Comp.}\ }\textbf {\bibinfo {volume} {25}},\ \bibinfo {pages}
  {671} (\bibinfo {year} {2013})}\BibitemShut {NoStop}%
\bibitem [{\citenamefont {Lu}\ \emph {et~al.}(2018)\citenamefont {Lu},
  \citenamefont {Hunt},\ and\ \citenamefont {Ott}}]{LHO:2018}%
  \BibitemOpen
  \bibfield  {author} {\bibinfo {author} {\bibfnamefont {Z.}~\bibnamefont
  {Lu}}, \bibinfo {author} {\bibfnamefont {B.}~\bibnamefont {Hunt}}, \ and\
  \bibinfo {author} {\bibfnamefont {E.}~\bibnamefont {Ott}},\ }\bibfield
  {title} {\enquote {\bibinfo {title} {Attractor reconstruction by machine
  learning},}\ }\href@noop {} {\bibfield  {journal} {\bibinfo  {journal}
  {Chaos}\ }\textbf {\bibinfo {volume} {28}},\ \bibinfo {pages} {061104}
  (\bibinfo {year} {2018})}\BibitemShut {NoStop}%
\bibitem [{\citenamefont {Chen}\ \emph {et~al.}(2009)\citenamefont {Chen},
  \citenamefont {Qiu},\ and\ \citenamefont {Huang}}]{CQH:2009}%
  \BibitemOpen
  \bibfield  {author} {\bibinfo {author} {\bibfnamefont {L.}~\bibnamefont
  {Chen}}, \bibinfo {author} {\bibfnamefont {C.}~\bibnamefont {Qiu}}, \ and\
  \bibinfo {author} {\bibfnamefont {H.~B.}\ \bibnamefont {Huang}},\ }\bibfield
  {title} {\enquote {\bibinfo {title} {Synchronization with on-off coupling:
  Role of time scales in network dynamics},}\ }\href {\doibase
  10.1103/PhysRevE.79.045101} {\bibfield  {journal} {\bibinfo  {journal} {Phys.
  Rev. E}\ }\textbf {\bibinfo {volume} {79}},\ \bibinfo {pages} {045101}
  (\bibinfo {year} {2009})}\BibitemShut {NoStop}%
\bibitem [{\citenamefont {Li}\ \emph {et~al.}(2018)\citenamefont {Li},
  \citenamefont {Sun}, \citenamefont {Chen},\ and\ \citenamefont
  {Wang}}]{LSCW:2018}%
  \BibitemOpen
  \bibfield  {author} {\bibinfo {author} {\bibfnamefont {S.}~\bibnamefont
  {Li}}, \bibinfo {author} {\bibfnamefont {N.}~\bibnamefont {Sun}}, \bibinfo
  {author} {\bibfnamefont {L.}~\bibnamefont {Chen}}, \ and\ \bibinfo {author}
  {\bibfnamefont {X.}~\bibnamefont {Wang}},\ }\bibfield  {title} {\enquote
  {\bibinfo {title} {Network synchronization with periodic coupling},}\ }\href
  {\doibase 10.1103/PhysRevE.98.012304} {\bibfield  {journal} {\bibinfo
  {journal} {Phys. Rev. E}\ }\textbf {\bibinfo {volume} {98}},\ \bibinfo
  {pages} {012304} (\bibinfo {year} {2018})}\BibitemShut {NoStop}%
\bibitem [{\citenamefont {Aranson}\ and\ \citenamefont
  {Kramer}(2002)}]{AK:2002}%
  \BibitemOpen
  \bibfield  {author} {\bibinfo {author} {\bibfnamefont {I.~S.}\ \bibnamefont
  {Aranson}}\ and\ \bibinfo {author} {\bibfnamefont {L.}~\bibnamefont
  {Kramer}},\ }\bibfield  {title} {\enquote {\bibinfo {title} {The world of the
  complex {Ginzburg-Landau} equation},}\ }\href@noop {} {\bibfield  {journal}
  {\bibinfo  {journal} {Rev. Mod. Phys}\ }\textbf {\bibinfo {volume} {74}},\
  \bibinfo {pages} {99} (\bibinfo {year} {2002})}\BibitemShut {NoStop}%
\bibitem [{\citenamefont {Cross}\ and\ \citenamefont
  {Hohenberg}(1993)}]{CH:1993}%
  \BibitemOpen
  \bibfield  {author} {\bibinfo {author} {\bibfnamefont {M.~C.}\ \bibnamefont
  {Cross}}\ and\ \bibinfo {author} {\bibfnamefont {P.~C.}\ \bibnamefont
  {Hohenberg}},\ }\bibfield  {title} {\enquote {\bibinfo {title} {Pattern
  formation outside of equilibrium},}\ }\href {\doibase
  10.1103/RevModPhys.65.851} {\bibfield  {journal} {\bibinfo  {journal} {Rev.
  Mod. Phys.}\ }\textbf {\bibinfo {volume} {65}},\ \bibinfo {pages} {851}
  (\bibinfo {year} {1993})}\BibitemShut {NoStop}%
\bibitem [{\citenamefont {Kuramoto}(1984)}]{Kbook:1984}%
  \BibitemOpen
  \bibfield  {author} {\bibinfo {author} {\bibfnamefont {Y.}~\bibnamefont
  {Kuramoto}},\ }\href@noop {} {\emph {\bibinfo {title} {Chemical Oscillations,
  Waves and Turbulence}}}\ (\bibinfo  {publisher} {Springer},\ \bibinfo
  {address} {Berlin},\ \bibinfo {year} {1984})\BibitemShut {NoStop}%
\bibitem [{\citenamefont {Heagy}\ \emph {et~al.}(1995)\citenamefont {Heagy},
  \citenamefont {Pecora},\ and\ \citenamefont {Carroll}}]{HPC:1995}%
  \BibitemOpen
  \bibfield  {author} {\bibinfo {author} {\bibfnamefont {J.~F.}\ \bibnamefont
  {Heagy}}, \bibinfo {author} {\bibfnamefont {L.~M.}\ \bibnamefont {Pecora}}, \
  and\ \bibinfo {author} {\bibfnamefont {T.~L.}\ \bibnamefont {Carroll}},\
  }\bibfield  {title} {\enquote {\bibinfo {title} {Short wavelength
  bifurcations and size instabilities in coupled oscillator systems},}\ }\href
  {\doibase 10.1103/PhysRevLett.74.4185} {\bibfield  {journal} {\bibinfo
  {journal} {Phys. Rev. Lett.}\ }\textbf {\bibinfo {volume} {74}},\ \bibinfo
  {pages} {4185} (\bibinfo {year} {1995})}\BibitemShut {NoStop}%
\bibitem [{\citenamefont {Pecora}\ and\ \citenamefont
  {Carroll}(1998)}]{PC:1998}%
  \BibitemOpen
  \bibfield  {author} {\bibinfo {author} {\bibfnamefont {L.~M.}\ \bibnamefont
  {Pecora}}\ and\ \bibinfo {author} {\bibfnamefont {T.~L.}\ \bibnamefont
  {Carroll}},\ }\bibfield  {title} {\enquote {\bibinfo {title} {Master
  stability functions for synchronized coupled systems},}\ }\href {\doibase
  10.1103/PhysRevLett.80.2109} {\bibfield  {journal} {\bibinfo  {journal}
  {Phys. Rev. Lett.}\ }\textbf {\bibinfo {volume} {80}},\ \bibinfo {pages}
  {2109} (\bibinfo {year} {1998})}\BibitemShut {NoStop}%
\bibitem [{\citenamefont {R\"{o}ssler}(1976)}]{Rossler:1976}%
  \BibitemOpen
  \bibfield  {author} {\bibinfo {author} {\bibfnamefont {O.~E.}\ \bibnamefont
  {R\"{o}ssler}},\ }\bibfield  {title} {\enquote {\bibinfo {title} {Equation
  for continuous chaos},}\ }\href@noop {} {\bibfield  {journal} {\bibinfo
  {journal} {Phys. Lett. A}\ }\textbf {\bibinfo {volume} {57}},\ \bibinfo
  {pages} {397} (\bibinfo {year} {1976})}\BibitemShut {NoStop}%
\bibitem [{\citenamefont {Lorenz}(1963)}]{Lorenz:1963}%
  \BibitemOpen
  \bibfield  {author} {\bibinfo {author} {\bibfnamefont {E.~N.}\ \bibnamefont
  {Lorenz}},\ }\bibfield  {title} {\enquote {\bibinfo {title} {Deterministic
  nonperiodic flow},}\ }\href@noop {} {\bibfield  {journal} {\bibinfo
  {journal} {J. Atmos. Sci.}\ }\textbf {\bibinfo {volume} {20}},\ \bibinfo
  {pages} {130} (\bibinfo {year} {1963})}\BibitemShut {NoStop}%
\bibitem [{\citenamefont {Hindmarsh}\ and\ \citenamefont
  {Rose}(1984)}]{HR:1984}%
  \BibitemOpen
  \bibfield  {author} {\bibinfo {author} {\bibfnamefont {J.~L.}\ \bibnamefont
  {Hindmarsh}}\ and\ \bibinfo {author} {\bibfnamefont {R.~M.}\ \bibnamefont
  {Rose}},\ }\bibfield  {title} {\enquote {\bibinfo {title} {A model of
  neuronal bursting using three coupled first order differential equations},}\
  }\href@noop {} {\bibfield  {journal} {\bibinfo  {journal} {Proc. Roy. Soc.
  London Ser. B Biol. Sci.}\ }\textbf {\bibinfo {volume} {221}},\ \bibinfo
  {pages} {87} (\bibinfo {year} {1984})}\BibitemShut {NoStop}%
\bibitem [{\citenamefont {Blasius}\ \emph {et~al.}(1999)\citenamefont
  {Blasius}, \citenamefont {Huppert},\ and\ \citenamefont {Stone}}]{BHS:1999}%
  \BibitemOpen
  \bibfield  {author} {\bibinfo {author} {\bibfnamefont {B.}~\bibnamefont
  {Blasius}}, \bibinfo {author} {\bibfnamefont {A.}~\bibnamefont {Huppert}}, \
  and\ \bibinfo {author} {\bibfnamefont {L.}~\bibnamefont {Stone}},\ }\bibfield
   {title} {\enquote {\bibinfo {title} {Complex dynamics and phase
  synchronization in spatially extended ecological systems},}\ }\href@noop {}
  {\bibfield  {journal} {\bibinfo  {journal} {Nature}\ }\textbf {\bibinfo
  {volume} {399}},\ \bibinfo {pages} {354} (\bibinfo {year}
  {1999})}\BibitemShut {NoStop}%
\bibitem [{\citenamefont {Patil}\ \emph {et~al.}(2001)\citenamefont {Patil},
  \citenamefont {Hunt}, \citenamefont {Kalnay}, \citenamefont {Yorke},\ and\
  \citenamefont {Ott}}]{PHKYO:2001}%
  \BibitemOpen
  \bibfield  {author} {\bibinfo {author} {\bibfnamefont {D.~J.}\ \bibnamefont
  {Patil}}, \bibinfo {author} {\bibfnamefont {B.~R.}\ \bibnamefont {Hunt}},
  \bibinfo {author} {\bibfnamefont {E.}~\bibnamefont {Kalnay}}, \bibinfo
  {author} {\bibfnamefont {J.~A.}\ \bibnamefont {Yorke}}, \ and\ \bibinfo
  {author} {\bibfnamefont {E.}~\bibnamefont {Ott}},\ }\bibfield  {title}
  {\enquote {\bibinfo {title} {Local low dimensionality of atmospheric
  dynamics},}\ }\href {\doibase 10.1103/PhysRevLett.86.5878} {\bibfield
  {journal} {\bibinfo  {journal} {Phys. Rev. Lett.}\ }\textbf {\bibinfo
  {volume} {86}},\ \bibinfo {pages} {5878} (\bibinfo {year}
  {2001})}\BibitemShut {NoStop}%
\bibitem [{\citenamefont {Ott}\ \emph {et~al.}(2004)\citenamefont {Ott},
  \citenamefont {Hunt}, \citenamefont {Szunyogh}, \citenamefont {Zimin},
  \citenamefont {Kostelich}, \citenamefont {Corazza}, \citenamefont {Kalnay},
  \citenamefont {Patil},\ and\ \citenamefont {Yorke}}]{Ottetal:2004}%
  \BibitemOpen
  \bibfield  {author} {\bibinfo {author} {\bibfnamefont {E.}~\bibnamefont
  {Ott}}, \bibinfo {author} {\bibfnamefont {B.}~\bibnamefont {Hunt}}, \bibinfo
  {author} {\bibfnamefont {I.}~\bibnamefont {Szunyogh}}, \bibinfo {author}
  {\bibfnamefont {A.~V.}\ \bibnamefont {Zimin}}, \bibinfo {author}
  {\bibfnamefont {E.~J.}\ \bibnamefont {Kostelich}}, \bibinfo {author}
  {\bibfnamefont {M.}~\bibnamefont {Corazza}}, \bibinfo {author} {\bibfnamefont
  {E.}~\bibnamefont {Kalnay}}, \bibinfo {author} {\bibfnamefont {D.~J.}\
  \bibnamefont {Patil}}, \ and\ \bibinfo {author} {\bibfnamefont {J.~A.}\
  \bibnamefont {Yorke}},\ }\bibfield  {title} {\enquote {\bibinfo {title} {A
  local ensemble {Kalman} filter for atmospheric data assimilation},}\
  }\href@noop {} {\bibfield  {journal} {\bibinfo  {journal} {Tellus}\ }\textbf
  {\bibinfo {volume} {56A}},\ \bibinfo {pages} {415} (\bibinfo {year}
  {2004})}\BibitemShut {NoStop}%
\bibitem [{\citenamefont {Szunyogh}\ \emph {et~al.}(2005)\citenamefont
  {Szunyogh}, \citenamefont {Kostelich}, \citenamefont {Gyarmati},
  \citenamefont {Patil}, \citenamefont {Hunt}, \citenamefont {Kalnay},
  \citenamefont {Ott},\ and\ \citenamefont {Yorke}}]{SKGPHKOY:2005}%
  \BibitemOpen
  \bibfield  {author} {\bibinfo {author} {\bibfnamefont {I.}~\bibnamefont
  {Szunyogh}}, \bibinfo {author} {\bibfnamefont {E.}~\bibnamefont {Kostelich}},
  \bibinfo {author} {\bibfnamefont {G.}~\bibnamefont {Gyarmati}}, \bibinfo
  {author} {\bibfnamefont {D.~J.}\ \bibnamefont {Patil}}, \bibinfo {author}
  {\bibfnamefont {B.~R.}\ \bibnamefont {Hunt}}, \bibinfo {author}
  {\bibfnamefont {E.}~\bibnamefont {Kalnay}}, \bibinfo {author} {\bibfnamefont
  {E.}~\bibnamefont {Ott}}, \ and\ \bibinfo {author} {\bibfnamefont {J.~A.}\
  \bibnamefont {Yorke}},\ }\bibfield  {title} {\enquote {\bibinfo {title}
  {Assessing a local ensemble {Kalman} filter: Perfect model experiments with
  the national centers for environmental prediction global model},}\
  }\href@noop {} {\bibfield  {journal} {\bibinfo  {journal} {Tellus}\ }\textbf
  {\bibinfo {volume} {57A}},\ \bibinfo {pages} {528} (\bibinfo {year}
  {2005})}\BibitemShut {NoStop}%
\bibitem [{\citenamefont {Brajard}\ \emph {et~al.}(2019)\citenamefont
  {Brajard}, \citenamefont {Carrassi}, \citenamefont {Bocquet},\ and\
  \citenamefont {Bertino}}]{JAML:2019}%
  \BibitemOpen
  \bibfield  {author} {\bibinfo {author} {\bibfnamefont {J.}~\bibnamefont
  {Brajard}}, \bibinfo {author} {\bibfnamefont {A.}~\bibnamefont {Carrassi}},
  \bibinfo {author} {\bibfnamefont {M.}~\bibnamefont {Bocquet}}, \ and\
  \bibinfo {author} {\bibfnamefont {L.}~\bibnamefont {Bertino}},\ }\bibfield
  {title} {\enquote {\bibinfo {title} {Combining data assimilation and machine
  learning to emulate a dynamical model from sparse and noisy observations: a
  case study with the {Lorenz} 96 model},}\ }\href@noop {} {\bibfield
  {journal} {\bibinfo  {journal} {Geosci. Model Dev. Discuss.}\ } (\bibinfo
  {year} {2019})}\BibitemShut {NoStop}%
\bibitem [{\citenamefont {Broomhead}\ and\ \citenamefont
  {Lowe}(1988)}]{BL:1988}%
  \BibitemOpen
  \bibfield  {author} {\bibinfo {author} {\bibfnamefont {D.~S.}\ \bibnamefont
  {Broomhead}}\ and\ \bibinfo {author} {\bibfnamefont {D.}~\bibnamefont
  {Lowe}},\ }\bibfield  {title} {\enquote {\bibinfo {title} {Multivariable
  functional interpolation and adaptive networks},}\ }\href@noop {} {\bibfield
  {journal} {\bibinfo  {journal} {Complex Syst.}\ }\textbf {\bibinfo {volume}
  {2}},\ \bibinfo {pages} {321} (\bibinfo {year} {1988})}\BibitemShut {NoStop}%
\bibitem [{\citenamefont {Poggio}\ and\ \citenamefont
  {Girosi}(1990)}]{PG:1990}%
  \BibitemOpen
  \bibfield  {author} {\bibinfo {author} {\bibfnamefont {T.}~\bibnamefont
  {Poggio}}\ and\ \bibinfo {author} {\bibfnamefont {F.}~\bibnamefont
  {Girosi}},\ }\bibfield  {title} {\enquote {\bibinfo {title} {Networks for
  approximation and learning},}\ }\href@noop {} {\bibfield  {journal} {\bibinfo
   {journal} {Proc. IEEE}\ }\textbf {\bibinfo {volume} {78}},\ \bibinfo {pages}
  {1484} (\bibinfo {year} {1990})}\BibitemShut {NoStop}%
\bibitem [{\citenamefont {Gholipour}\ \emph {et~al.}(2006)\citenamefont
  {Gholipour}, \citenamefont {Araabi},\ and\ \citenamefont {Lucas}}]{GAL:2006}%
  \BibitemOpen
  \bibfield  {author} {\bibinfo {author} {\bibfnamefont {A.}~\bibnamefont
  {Gholipour}}, \bibinfo {author} {\bibfnamefont {B.~N.}\ \bibnamefont
  {Araabi}}, \ and\ \bibinfo {author} {\bibfnamefont {C.}~\bibnamefont
  {Lucas}},\ }\bibfield  {title} {\enquote {\bibinfo {title} {Predicting
  chaotic time series using neural and neurofuzzy models: A comparative
  study},}\ }\href@noop {} {\bibfield  {journal} {\bibinfo  {journal} {Neural
  Process. Lett.}\ }\textbf {\bibinfo {volume} {24}},\ \bibinfo {pages} {217}
  (\bibinfo {year} {2006})}\BibitemShut {NoStop}%
\bibitem [{\citenamefont {Chen}\ and\ \citenamefont {Han}(2013)}]{CH:2013}%
  \BibitemOpen
  \bibfield  {author} {\bibinfo {author} {\bibfnamefont {D.}~\bibnamefont
  {Chen}}\ and\ \bibinfo {author} {\bibfnamefont {W.}~\bibnamefont {Han}},\
  }\bibfield  {title} {\enquote {\bibinfo {title} {Prediction of multivariate
  chaotic time series via radial basis function neural network},}\ }\href@noop
  {} {\bibfield  {journal} {\bibinfo  {journal} {Complexity}\ }\textbf
  {\bibinfo {volume} {18}},\ \bibinfo {pages} {55} (\bibinfo {year}
  {2013})}\BibitemShut {NoStop}%
\bibitem [{\citenamefont {Rafsanjani}\ and\ \citenamefont
  {Samareh}(2016)}]{RS:2016}%
  \BibitemOpen
  \bibfield  {author} {\bibinfo {author} {\bibfnamefont {M.~K.}\ \bibnamefont
  {Rafsanjani}}\ and\ \bibinfo {author} {\bibfnamefont {M.}~\bibnamefont
  {Samareh}},\ }\bibfield  {title} {\enquote {\bibinfo {title} {Chaotic time
  series prediction by artificial neural networks},}\ }\href@noop {} {\bibfield
   {journal} {\bibinfo  {journal} {J. Comp. Methods Sci. Eng.}\ }\textbf
  {\bibinfo {volume} {16}},\ \bibinfo {pages} {599} (\bibinfo {year}
  {2016})}\BibitemShut {NoStop}%
\bibitem [{\citenamefont {Nguyen}\ and\ \citenamefont {Duong}(2018)}]{ND:2018}%
  \BibitemOpen
  \bibfield  {author} {\bibinfo {author} {\bibfnamefont {V.~T.}\ \bibnamefont
  {Nguyen}}\ and\ \bibinfo {author} {\bibfnamefont {T.~A.}\ \bibnamefont
  {Duong}},\ }\bibfield  {title} {\enquote {\bibinfo {title} {Chaotic time
  series prediction using radial basis function networks},}\ }in\ \href@noop {}
  {\emph {\bibinfo {booktitle} {2018 4th International Conference on Green
  Technology and Sustainable Development (GTSD)}}}\ (\bibinfo {year}
  {2018})\BibitemShut {NoStop}%
\bibitem [{\citenamefont {Stefa\ifmmode~\acute{n}\else \'{n}\fi{}ski}\ \emph
  {et~al.}(2007)\citenamefont {Stefa\ifmmode~\acute{n}\else \'{n}\fi{}ski},
  \citenamefont {Perlikowski},\ and\ \citenamefont {Kapitaniak}}]{SPK:2007}%
  \BibitemOpen
  \bibfield  {author} {\bibinfo {author} {\bibfnamefont {A.}~\bibnamefont
  {Stefa\ifmmode~\acute{n}\else \'{n}\fi{}ski}}, \bibinfo {author}
  {\bibfnamefont {P.}~\bibnamefont {Perlikowski}}, \ and\ \bibinfo {author}
  {\bibfnamefont {T.}~\bibnamefont {Kapitaniak}},\ }\bibfield  {title}
  {\enquote {\bibinfo {title} {Ragged synchronizability of coupled
  oscillators},}\ }\href {\doibase 10.1103/PhysRevE.75.016210} {\bibfield
  {journal} {\bibinfo  {journal} {Phys. Rev. E}\ }\textbf {\bibinfo {volume}
  {75}},\ \bibinfo {pages} {016210} (\bibinfo {year} {2007})}\BibitemShut
  {NoStop}%
\bibitem [{\citenamefont {Perlikowski}\ \emph {et~al.}(2008)\citenamefont
  {Perlikowski}, \citenamefont {Jagiello}, \citenamefont {Stefanski},\ and\
  \citenamefont {Kapitaniak}}]{PJSK:2008}%
  \BibitemOpen
  \bibfield  {author} {\bibinfo {author} {\bibfnamefont {P.}~\bibnamefont
  {Perlikowski}}, \bibinfo {author} {\bibfnamefont {B.}~\bibnamefont
  {Jagiello}}, \bibinfo {author} {\bibfnamefont {A.}~\bibnamefont {Stefanski}},
  \ and\ \bibinfo {author} {\bibfnamefont {T.}~\bibnamefont {Kapitaniak}},\
  }\bibfield  {title} {\enquote {\bibinfo {title} {Experimental observation of
  ragged synchronizability},}\ }\href {\doibase 10.1103/PhysRevE.78.017203}
  {\bibfield  {journal} {\bibinfo  {journal} {Phys. Rev. E}\ }\textbf {\bibinfo
  {volume} {78}},\ \bibinfo {pages} {017203} (\bibinfo {year}
  {2008})}\BibitemShut {NoStop}%
\end{thebibliography}

%
\end{document}